\documentclass[nohyperref]{article}

\usepackage{microtype}
\usepackage{graphicx}
\usepackage{subfigure}
\usepackage{booktabs} % for professional tables

\usepackage{hyperref}

\usepackage[accepted]{icml2022}

\usepackage{amsmath}
\usepackage{amssymb}
\usepackage{mathtools}
\usepackage{amsthm}

\usepackage{bm, graphicx}

\def\K{\bm K}
\def\x{\bm x}
\def\bPsi{\bm\bPsi}
\def\y{\bm y}

\usepackage[capitalize,noabbrev]{cleveref}

\theoremstyle{plain}

\theoremstyle{definition}

\theoremstyle{remark}

\usepackage[textsize=tiny]{todonotes}

\icmltitlerunning{A Theory of Kernel Alignment During Neural Network Training}

\begin{document}

\twocolumn[
\icmltitle{A Theory of Neural Tangent Kernel Alignment and Its Influence on Training}

% It is OKAY to include author information, even for blind
% submissions: the style file will automatically remove it for you
% unless you've provided the [accepted] option to the icml2022
% package.

% List of affiliations: The first argument should be a (short)
% identifier you will use later to specify author affiliations
% Academic affiliations should list Department, University, City, Region, Country
% Industry affiliations should list Company, City, Region, Country

% You can specify symbols, otherwise they are numbered in order.
% Ideally, you should not use this facility. Affiliations will be numbered
% in order of appearance and this is the preferred way.
\icmlsetsymbol{equal}{*}

\begin{icmlauthorlist}
\icmlauthor{Haozhe Shan}{equal,cbs}
\icmlauthor{Blake Bordelon}{equal,seas}
\end{icmlauthorlist}

\icmlaffiliation{cbs}{Center for Brain Science, Harvard University, Cambridge, MA, United States}
\icmlaffiliation{seas}{School of Engineering and Applied Sciences, Harvard University, Cambridge, MA, United States}

\icmlcorrespondingauthor{Haozhe Shan}{hshan@g.harvard.edu}
\icmlcorrespondingauthor{Blake Bordelon}{blake\_bordelon@g.harvard.edu}

% You may provide any keywords that you
% find helpful for describing your paper; these are used to populate
% the "keywords" metadata in the PDF but will not be shown in the document

\vskip 0.3in
]

% this must go after the closing bracket ] following \twocolumn[ ...

% This command actually creates the footnote in the first column
% listing the affiliations and the copyright notice.
% The command takes one argument, which is text to display at the start of the footnote.
% The \icmlEqualContribution command is standard text for equal contribution.
% Remove it (just {}) if you do not need this facility.

%\printAffiliationsAndNotice{}  % leave blank if no need to mention equal contribution
\printAffiliationsAndNotice{\icmlEqualContribution} % otherwise use the standard text.

\begin{abstract}
The training dynamics and generalization properties of neural networks (NN) can be precisely characterized in function space via the neural tangent kernel (NTK). Structural changes to the NTK during training reflect feature learning and underlie the superior performance of networks outside of the static kernel regime. In this work, we seek to theoretically understand kernel alignment, a prominent and ubiquitous structural change that aligns the NTK with the target function. We first study a toy model of kernel evolution in which the NTK evolves to accelerate training and show that alignment naturally emerges from this demand. We then study alignment mechanism in deep linear networks and two layer ReLU networks. These theories provide good qualitative descriptions of kernel alignment and specialization in practical networks and identify factors in network architecture and data structure that drive kernel alignment. In nonlinear networks with multiple outputs, we identify the phenomenon of kernel specialization, where the kernel function for each output head preferentially aligns to its own target function. Together, our results provide a mechanistic explanation of how kernel alignment emerges during NN training and a normative explanation of how it benefits training.
\end{abstract}

%%%%%%% TEXT BEGINS HERE
\section{Introduction}
Deep learning provides a flexible framework to solve difficult statistical inference problems across a variety of application areas \cite{lecun2015deep}. During optimization of the statistical objective, useful features are often extracted by the neural network (NN) as the weights in the network evolve. Though feature learning appears to be crucial to the success of neural networks on large-scale problems \cite{Geiger_2020,Ghorbani_2021, yang2021feature} (as well as enabling transfer learning on new, related problems), the precise way that neural network features evolve to benefit learning is not well understood theoretically.

One powerful framework for studying feature learning in NNs is through the neural tangent kernel (NTK) \cite{jacot2020neural, Lee_2020}. This framework grew out of the observation that in the limit of large widths and small learning rates, NNs with certain parameterization behave like linear models in their parameters. In this case, NN training is equivalent to kernel gradient descent (KGD) with a static neural tangent kernel. While this limit allows a precise characterization of training and generalization dynamics to be obtained \cite{bietti2019inductive, yang2020finegrained, bordelon2020spectrum, bahri2021explaining}, the time stationarity of the kernel indicates that feature learning does not occur. In practical NNs, however, widths are finite and the NTK evolves during training \cite{dyer2019asymptotics, Geiger_2020}. In this work, we consider \textit{feature learning in NNs as the evolution of the NTK during training}. 

Recent empirical works have identified that a ubiquitous feature of the NTK evolution \cite{baratin_neural_feature_align, fort2020deep,Geiger_2020, atanasov2021neural, paccolat2021geometric} in practical settings is that the kernel aligns with the target function over time, a phenomenon we hereafter refer to as ``kernel alignment". The prominence and ubiquity of kernel alignment suggest that it may play an important role in NN feature learning. In particular, it has been speculated that this underlies the superior generalization performance of practical NNs, when compared to their infinite-width counterparts \cite{baratin_neural_feature_align, fort2020deep,Geiger_2020, paccolat2021geometric}. In this work, we aim to provide a theoretical understanding of the dynamics of kernel alignment during NN training and how it affects learning. Specifically, our main contributions are
\begin{itemize}
\setlength\itemsep{0.1pt}
    \item We demonstrate that kernel alignment accelerates training by showing that a kernel aligning over time accelerates convergence of the training loss. In particular, we study a toy model of kernel evolution where the NTK features explicitly evolve towards the direction that accelerates training and find that kernel alignment occurs as a result. The strength of the acceleration is controlled by a single parameter, the feature learning rate ($\gamma$), which determines not only the acceleration in training but also the final alignment of the NTK. 
    \item We provide an analytical theory of how kernel alignment emerges during gradient descent learning in deep linear networks(Sec.\ref{sec: mechanisms of KA in linear}) and an approximate treatment of two layer ReLU networks (Sec.\ref{sec:KA in ReLU network}). The theory captures key qualitative features of kernel alignment in real-world networks and makes the novel prediction that kernel alignment is stronger in deeper networks, which we validate numerically in ReLU networks. 
    \item We report the novel empirical finding of  ``kernel specialization" (Sec.\ref{subsec:specialization}), which occurs in NNs with multiple output heads (e.g., those for multiclass classification). For these NNs, the NTK has different components for each pair of output heads. We found that the diagonal components corresponding to each head becomes aligned with its specific target function. Our theory shows how this emerges from the interaction between network architecture and data structure.
\end{itemize}

\section{Related Work}
Characterizing and understanding the time evolution of the NTK for during NN training has been the subject of considerable interest in the deep learning theory community. This is motivated by the empirical finding that the NTK evolution underlies a significant performance gap between practical NNs and infinite-width NNs \cite{lee2020finite, Geiger_2020}. General expressions for the leading order corrections to NN dynamics for finite width networks with NTK parameterization can be characterized within the framework of perturbation theory through truncation of an infinite set of ODEs known as the neural tangent hierarchy \cite{dyer2019asymptotics, Huang_neural_tangent_hierarchy, aitken2020asymptotics, roberts2021principles}. This leading order truncation contains corrections which scale as $1/\text{width}$ and depends on data through the targets and a third and fourth order tensor of the inputs. The alternative mean-field infinite-width scaling reveal finite evolution of hidden features which depend on the supervised training signal which persist even at infinite width \cite{Mei2019MeanfieldTO, sirignano2020mean,yang2021feature, nguyen2019mean,phan_minh_rigorous}. These results, however, generally involve nonlinear partial differential equations for the parameter or feature distributions. 

Instead of characterizing evolution of the entire kernel, recent empirical and theoretical works have focused on describing and analyzing specific features of the evolution. In particular, kernel alignment has emerged as a prominent and ubiquitous phenomenon in this type of analysis. \cite{fort2020deep} find an initial transient period of kernel evolution followed by a period where the kernel can be approximated as static. \cite{baratin_neural_feature_align} showed that the early dynamics align NTK eigenvectors with task relevant directions. \cite{atanasov2021neural} provide conditions for a ``silent alignment" effect in networks with small initialization in which the NTK aligns to the task relevant subspace before scale growth, giving network output which is a kernel regression solution with the final, rather than initial, NTK. 

% Experimental studies in the large width regime have indicated that finite width CNNs enjoy an advantage on computer vision problems compared to their kernel counterparts but that the generalization gap between finite and infinite width networks is heavily dependent on architecture and data structure \cite{Geiger_2020, lee2020finite}. 

\begin{figure}
  \centering
  \includegraphics[width=\linewidth]{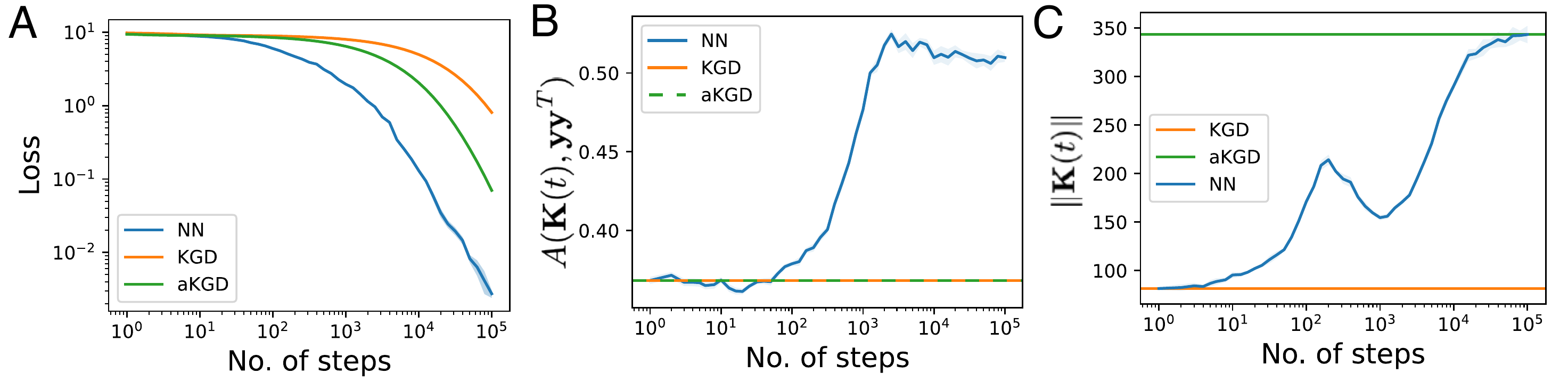}
  \caption{\textbf{Feature evolution alters the structure of the NTK and accelerates learning.}\\ \textbf{A} The training loss for a two-layer $N=500$ MLP when trained on a subset of MNIST (NN) is compared to kernel gradient descent with the initial kernel (KGD) and the initial kernel using a rescaled learning rate to account for the difference in norm of the NN's NTK and the initial NTK (aKGD). We see that even the optimistically rescaling of the learning rate by the final NN's NTK norm does not account for the gap in the loss. \textbf{B} The norm of the kernel increases non-monotonically throughout training. \textbf{C} The alignment between the NTK and the task kernel throughout training increases to an asymptote for the neural network but remains constant for the static kernel dynamics (KGD, aKGD). Average and standard deviation over five different initializtions are plotted.}
  \label{fig:kgd_vs_nn}
\end{figure}

\section{Preliminaries: NTK and Kernel Alignment}
In this section, we briefly review the NTK definition \cite{jacot2020kernel} and give a precise definition of kernel alignment during NN training. For simplicity, we will first consider NNs with scalar output functions and will extend our discussion to multiple class outputs in a later section. Let $f(\x,\bm\theta)$ represent the output of a neural network with parameters $\bm\theta$ and input vector $\x$. We optimize the parameters $\bm\theta$ with gradient flow on a loss function $\mathcal{L}=\sum_\mu \ell(f(\x^\mu,\bm\theta), y^\mu)$ with $P$ examples $\mathcal{D}= \{(\x^\mu, y^\mu)\}_{\mu=1}^P$ in the training set. Throughout this work, we assume batch gradient descent dynamics, which give
\begin{equation}
    \frac{d \bm\theta}{dt}= -\eta \sum_{\mu=1}^P \frac{\partial f(\x^\mu,\bm\theta)}{\partial \bm\theta} \frac{\partial \ell(f(\x^\mu,\bm\theta), y^\mu) }{\partial f(\x^\mu,\bm\theta)}
    \label{eq:training setting}
\end{equation}
where $\eta$ is the learning rate. Rather than studying the dynamics of the parameters $\bm\theta$, the NTK formulation focuses on dynamics of the network output,
\begin{equation}\label{eq:exact_dynamics_ntk}
    \frac{d f(\x^\mu,\bm\theta)}{dt} = - \eta \sum_{\nu} K(\x^\mu,\x^{\nu};\bm \theta) \frac{\partial \ell(f(\x^\mu,\bm\theta), y^\mu) }{\partial f(\x^\nu,\bm\theta)}
\end{equation}
where the $K$ represents the NTK $K(\x,\x' ;\bm \theta) = \frac{\partial f(\x,\bm\theta)}{\partial \bm\theta} \cdot \frac{\partial f(\x',\bm\theta)}{\partial \bm\theta}$. On the training set, it is completely described by the gram matrix $\bm K(\bm\theta) \in \mathbb{R}^{P\times P}$. 

On the training set, the NTK evolution, including kernel alignment, is completely reflected by the dynamics of this matrix over time. Importantly, this evolution entirely characterizes the difference from the infinite-width limit, where the NTK is static, and the regime of practical NNs. As a simple demonstration of this phenomenon, we trained MLPs with two hidden layers with 500 hidden ReLU units per layer to do odd-even binary classification of MNIST dataset. We compared the dynamics of its training loss (``NN") against the expected loss dynamics if the NTK were frozen at its initial value (``KGD") and found that the NN training is significantly faster(Fig.\ref{fig:kgd_vs_nn}\textbf{A}). Importantly, while the NN $\bm K(\bm\theta)$ increases in its Frobenius norm(Fig.\ref{fig:kgd_vs_nn}\textbf{B}), as reported previously \cite{baratin_neural_feature_align}, this alone does not account for the acceleration of learning. To see this, we simulated the training loss dynamics of KGD using the initial NTK but scaled up to match the amplitude of the final NN NTK and found that it still could not match NN performance (Fig.\ref{fig:kgd_vs_nn}\textbf{A}). Therefore, understanding changes to the structure of $\bm{K}$, rather than simply its amplitude, is essential.

To quantify kernel alignment, it is useful to introduce the \textit{kernel alignment} metric \cite{cortes}.
\begin{equation}
    A(t) = \frac{\left< \bm y \bm y^\top,  \bm K(\bm\theta) \right>_F}{||\K(\bm\theta)||_F ||\bm \y \bm y^\top ||_F}=  \frac{\bm y^\top \bm K(\bm\theta) \bm y}{||\K(\bm\theta)||_F ||\bm \y ||^2}.
\end{equation}
Indeed, $A(t)$ increases substantially in our experiment before stabilizing(Fig.\ref{fig:kgd_vs_nn}\textbf{C}). 

\section{How Kernel Alignment Influences Learning}

\subsection{Kernel Alignment and Optimal Feature Evolution}
\label{sec:optimal feature evolution}
We first investigate how kernel alignment influences learning. In particular, we would like to understand if and how kernel alignment is responsible for the accelerated decrease of training loss seen in Fig.\ref{fig:kgd_vs_nn}. To do so, we first step away from NNs for a moment and consider the general case of KGD with an evolving kernel $\K(t) = \bm\Psi(t)^\top \bm\Psi(t)$. The very important issue of \textit{how} kernel alignment emerges during NN training is addressed in Sec.\ref{sec: mechanisms of KA in linear} and Sec.\ref{sec:KA in ReLU network}. 

The matrix $\bm\Psi(t) \in \mathbb{R}^{P \times Q}$, where $Q$ is the number of parameters in the model, defines the feature map used by the kernel. Assuming a mean squared error loss, the loss dynamics of KGD is given by
\begin{equation}
    \mathcal L_{t+1}(\bm\Psi) = ||(\bm I - \eta \bm\Psi_t^\top \bm\Psi_t) \bm\Delta_t ||^2,
\end{equation}
where $\bm\Delta_t = \bm f_t - \bm y \in \mathbb{R}^P$. 

In order to formulate a toy model of ``optimal" feature evolution for accelerating learning, we consider the case where the feature map is explicitly updated via gradient descent on $\mathcal{L}_{t+1}$ to accelerate learning, i.e.
\begin{equation}
    \bm\Psi_{t+1} = \bm\Psi_t - \gamma \frac{\partial L_{t+1}}{\partial \bm\Psi}|_{\bm\Psi_t}.
\end{equation}
Here, $\gamma$ is a scalar which we hereafter refer to as the ``feature learning rate". The fact that it is finite reflects the constraint that the feature map cannot evolve infinitely fast during KGD. The limit $\gamma \rightarrow 0$ recovers the static kernel limit. In order to analyze what this optimal feature evolution (OFE) would look like, we go to the continuous time limit (Appendix \ref{app_optimal_evolution}) where
\begin{align}
    \dot{\bm\Delta}(t) &= - \eta \bm\Psi(t)^\top \bm\Psi(t) \bm\Delta(t) \nonumber\\
    \dot{\bm\Psi}(t) &= \gamma \eta \bm\Psi(t) \bm\Delta(t) \bm\Delta(t)^\top. 
    \label{eq:ofe equations}
\end{align}
In this limit, one finds that the matrix $\bm C = \gamma \bm\Delta(t) \bm\Delta(t)^\top + \bm \Psi(t)^\top \bm\Psi(t)$ is static in time (Appendix \ref{app_optimal_evolution}). Exploiting this fact and noting that the training loss $\lVert \bm\Delta(t) \rVert^2$ must eventually reach zero, one can identify the final kernel
\begin{equation}
    \K_\infty = \gamma \bm{y} \bm{y}^T + \bm{K}_0.
    \label{eq:ofe final kernel}
\end{equation}
Further, we can verify that increasing $\gamma$ is indeed beneficial to training dynamics by noting that  $\frac{1}{2} \frac{d}{dt} {|\bm\Delta|^2} = - \eta \bm \Delta^\top \left[ \K_0 + \gamma (\bm y \bm y^\top  - \bm\Delta \bm\Delta^\top) \right] \bm\Delta \leq -\eta \bm\Delta^\top \bm K_0 \bm\Delta$, which shows that positive $\gamma$ accelerates the convergence of the loss. 
The analysis of ``optimal feature evolution" here suggests that if one is to simultaneously optimize the features $\bm\Psi$ and the training errors $\bm\Delta$, the kernel will become aligned with the task functions, which is exactly kernel alignment. This expression bears remarkable resemblance the final NTK in the deep linear network case as we discuss in Sec.\ref{sec: mechanisms of KA in linear}(Eq.\ref{eq:linear network ntk}).

%\note{the next paragraph needs updating} 
While the OFE model does not necessarily capture the trajectory of the NTK, it suggests the heuristic that network depth $L$ acts like $\gamma$ to control how quickly the NTK evolves. We test this heuristic in the next section where we compare the training and alignment dynamics of real NNs with different feature learning rates. 
%To test this heuristic, we first trained two ReLU MLPs of different depths on an odd-vs-even MNIST task. Consistently with this heuristic, the deeper network learns faster (Fig.\ref{fig:ofe vs nn}, solid lines). We measured the effective $\gamma$ of these models using Eq.\ref{eq:ofe final kernel} and then simulated the differential equations in Eq.\ref{eq:ofe equations} (Fig.\ref{fig:ofe vs nn}, dashed lines). The results suggest that the kernel evolution during real NN training may be very close to the optimal one. \textcolor{red}{here i think additional empirical results can really help. also just do the OFE alignment trajectory vs NN trajectory}

\subsection{Enhancing Feature Evolution Through Rescaling}

As was explored in the prior works of \cite{Chizat, Geiger_2020}, the rate at which the NTK evolves can be altered through rescaling of the trainable network function: $g(\x) = \frac{1}{\gamma} f(\x)$ and learning rate $\eta = \eta_0 \gamma^2$. Letting, $L = |\bm\Delta(t)|^2 = |\y - \bm g(t)|^2$, gradient flow $\frac{d\bm\theta}{dt}= - \nabla_{\bm\theta} L$ gives $\frac{d L}{dt} = O_{\gamma}\left( 1 \right)$ and $\frac{d}{dt} \frac{\partial f}{\partial \bm\theta} = O_{\gamma}(\gamma)$ (see Appendix \ref{app:rescaling}). Recognizing that for NN dynamics, the relevant features are the parameter gradients $\bm\psi_\mu = \frac{\partial f_\mu }{\partial \bm \theta}$, we see that increasing $\gamma$ increases the relative rate at which the network gradients evolve compared to the loss at initialization, leading to faster kernel evolution. We illustrate that such rescaling indeed alters the kernel and training dynamics in a Wide ResNet \cite{zagoruyko2017wide} with fixed widening factor of $k = 3$ and block size $b = 2$ on a subset of $100$ CIFAR-10 images in Figure \ref{fig:WRN_alignment} (details in Appendix\ref{appd:wide resnet details}).  For $\gamma = 1$, the network is sufficiently wide that no significant feature evolution occurs, but at large $\gamma$, the NTK reaches a high final alignment value. Consistent with our OFE toy model, the NNs trained with highest $\gamma$ not only align more quickly but also reach a higher final alignment value. Further, increasing $\gamma$ accelerates training convergence due to superior alignment, again consistent with OFE. Estimating the feature learning rate for the OFE from the initial and final alignment values, we obtain qualitatively similar loss(Fig.\ref{fig:WRN_alignment}\textbf{A}vs.\textbf{C}) and alignment curves(Fig.\ref{fig:WRN_alignment}\textbf{B}vs.\textbf{D}). 
\begin{figure}
    \centering
    \includegraphics[width=\linewidth]{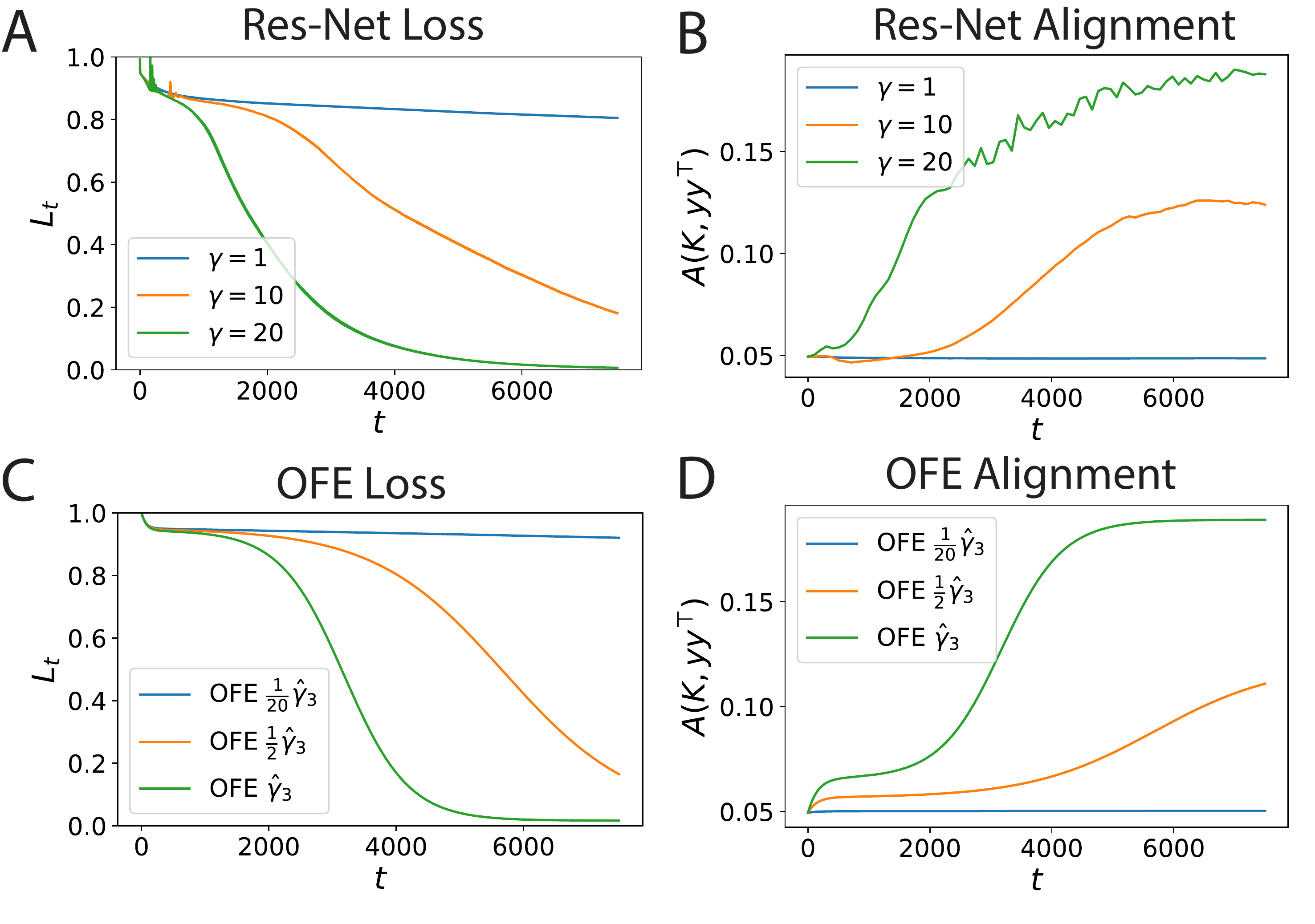}
    \caption{\textbf{Kernel and loss evolution in Wide Res-Net on CIFAR-10 subsample.} \textbf{A} Though the initial loss dynamics are similar for different $\gamma$, the drops more quickly after the NTK starts to aligning to the targets. \textbf{B} The alignment increases more quickly and reaches a higher final value for large $\gamma$, consistent with the toy OFE model. \textbf{C}-\textbf{D} The OFE model with $\K_0$ extracted as the initial NTK, and the feature learning rate $\hat{\gamma}_3$ estimated from the initial and final alignment of the green curve in panel B. Both loss and alignment dynamics are qualitatively similar to the real NN experiments.}
    \label{fig:WRN_alignment}
\end{figure}

\subsection{Kernel Specialization in Multiclass NNs}
\label{subsec:specialization}
So far, we have focused the analysis on NNs with a scalar output. We now consider NNs with multiclass outputs, which are common in practical applications. For networks with $C$ output nodes (e.g. $C$-class classification networks), the NTK $\bm{K}$ is a 4D tensor of dimensions $P\times P \times C \times C$. We define a matrix-valued "subkernel" as
\begin{equation}
    \bm K^{c,c'} \in \mathbb{R}^{P\times P}: (\bm K^{c,c'})_{\mu \mu'} = \nabla_{\bm \theta} f_c^\mu \cdot \nabla_{\bm \theta} f_{c'}^\nu 
    \label{eq:def of subkernel}
\end{equation}
We now consider a typical setting where the scalar activation of the $c$th output node, $f_c(\bm{x})$, is fitted to a separate target function (e.g., an indicator function of whether the input belongs to a certain class), $\bm y_c \in \mathbb{R}^P$. In addition, the loss function is decomposed into terms depending on individual nodes, e.g., $\mathcal L = C^{-1} P^{-1} \sum_{\mu=1}^P \sum_{c=1}^C (f_c(\bm{x}_\mu) - y_c^\mu)^2$. In this case, the dynamics of the loss for the $c$th output head only depend on $\bm{K^{c,c}}$.

At infinite network width,  $\bm K^{c,c'}=\bm{0}$ if $c\neq c'$ and all ``diagonal" subkernels, $\bm K^{c,c}$, are identical. This suggests that each output node evolves towards its own target function under KGD dynamnics governed by the same subkernel. Our analysis of the OFE suggests that to accelerate learning, it is best for each output node to evolve using a separate subkernel. Its own subkernel should learn features specific to its own target function, such that $\bm K^{c,c}$ becomes aligned with $\bm{y}_{c}$ but not $\bm{y}_{d \neq c}$. We call this phenomenon \textit{kernel specialization}.

%\begin{figure}[h]
%  \centering
%  \includegraphics[scale=0.5]{figures/ofe_vs_nn.pdf}
%  \caption{\textbf{Loss dynamics in NNs with different depths are well characterized by the optimal feature evolution model with different $\gamma$.}}
%  \label{fig:ofe vs nn}
%\end{figure}

Does kernel specialization occur in neural network learning? Previous empirical studies of multi-output networks studies the ``traced kernel", $\bm K_\text{tr} = C^{-1} \sum_{c=1}^C \bm K^{c,c} $, and found that it becomes aligned with all $C$ target functions \cite{baratin_neural_feature_align} during training. While this observation is consistent with kernel specialization, it can also arise from indiscriminate alignment where $\bm K^{c,c}$ becomes aligned with all $\{\bm{y}_{d}\}_{d=1,...,C}$. To see whether kernel specialization occurs, we first define a kernel specialization matrix (KSM), defined as
\begin{equation}
    \text{KSM}(c,d) \equiv \frac{A(\bm K^{c,c}, \bm y_{d} {\bm y_{d}}^T)}{C^{-1} \sum_{d'=1}^C A(\bm K^{d',d'}, \bm y^{d} {\bm y^{d}}^T)}.
    \label{eq:ksm definition}
\end{equation}
If kernel specialization occurs, KSM should be higher when $c=d$ (i.e. diagonal elements of the KSM). We computed the KSM for a two-layer $N=500$ MLP trained on 10-class classification of MNIST digits and found that diagonal elements are indeed higher than off-diagonal ones, demonstrating kernel specialization (Fig.\ref{fig:KSM} \textbf{A}, details in Appendix\ref{subsub: two layre relu on mnist}). The same qualitative result was replicated in a convolutional network trained on classification of CIFAR-10 images (Fig.\ref{fig:KSM} \textbf{B}, details in Appendix\ref{subsub: two layre relu on cifar10}; only two classes were used). 
\begin{figure}[h]
  \centering
  \includegraphics[width=\linewidth]{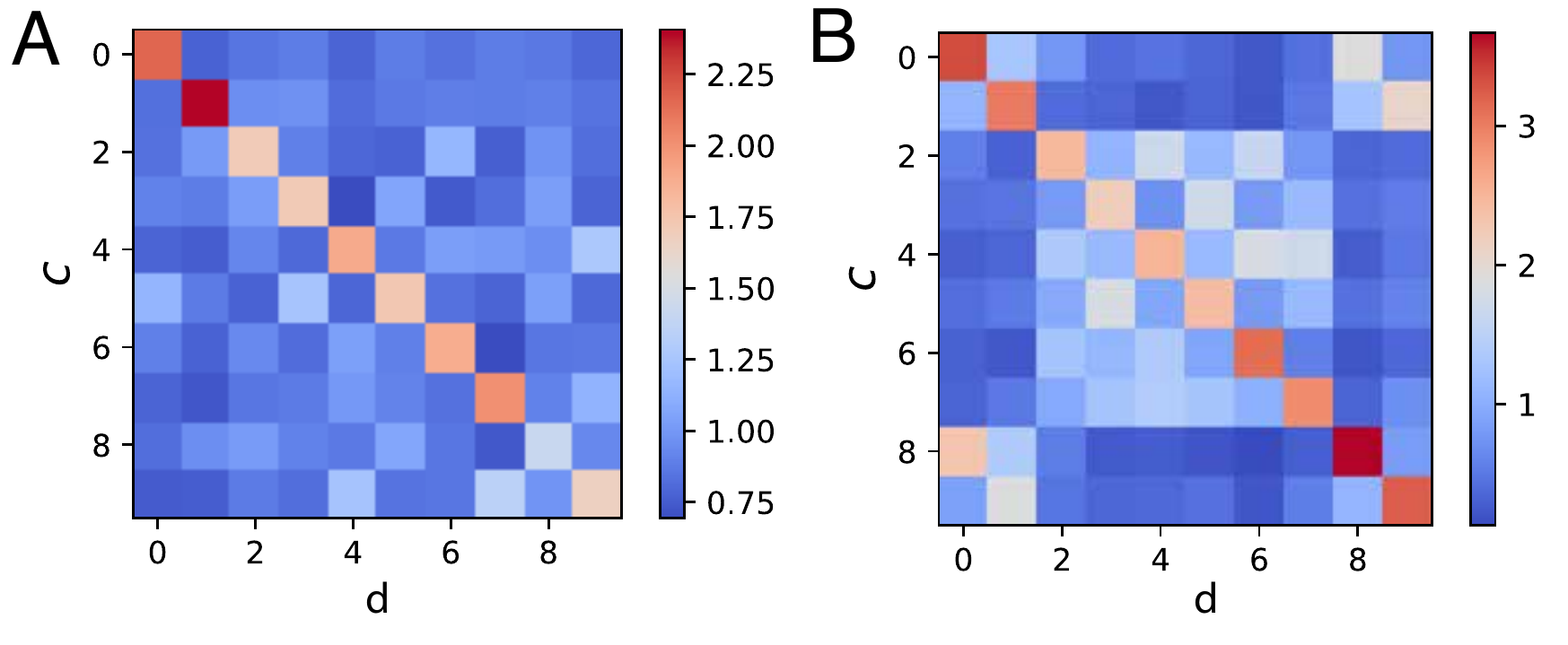}
  \caption{\textbf{NTK becomes specialized during training in NNs with multiclass outputs and nonlinear activation}. \textbf{A}. KSM (defined in Eq.\ref{eq:ksm definition}) of a two-layer $N=500$ ReLU MLP trained on 10-class classification of MNIST digits. \textbf{B}. KSM of a convolutional neural network trained on CIFAR-10. }
  \label{fig:KSM}
\end{figure}

\section{Mechanisms of Alignment in Linear NNs}
\label{sec: mechanisms of KA in linear}

Having shown that kernel alignment can indeed accelerate training, we now turn to the important question of how it \textit{mechanistically} arises from training NNs with gradient descent. In particular, we would like to identify important features of network architectures and data structures that give rise to kernel alignment. We begin by analyzing kernel alignment in deep linear networks in Sec.\ref{subsec:KA in linear networks} before moving on to two-layer ReLU networks in Sec.\ref{sec:KA in ReLU network}.

\subsection{Kernel Alignment in Deep Linear Networks}
\label{subsec:KA in linear networks}
Deep linear networks have supplied rich theoretical insights on NN training dynamics which generalize well to nonlinear counterparts \cite{advani2020high, cohen_arora, du2018algorithmic}. They are therefore a natural starting point for developing a theory of kernel alignment in NNs. First, we examined numerically whether kernel alignment occurs at all in linear networks. We trained a deep linear network with two hidden layers and a scalar output to learn a linear mapping $\mathbb{R}^N \rightarrow \mathbb{R}$, $y^\mu = \bm \beta \bm{x^\mu}$, where $\{\bm{x^\mu}\}$ are $P$ i.i.d. Gaussian vectors as the training set and $\bm \beta$ are the weights of a linear teacher. We found that the NTK becomes aligned with the target function in a similar fashion to the kernel alignment observed in our MLP experiment (Fig.\ref{fig:kgd_vs_nn}) and previous empirical work \cite{fort2020deep, Geiger_2020, baratin_neural_feature_align, atanasov2021neural}. This suggests that kernel alignment can indeed occur in deep linear networks.

\begin{figure}
  \centering
  \includegraphics[width=\linewidth]{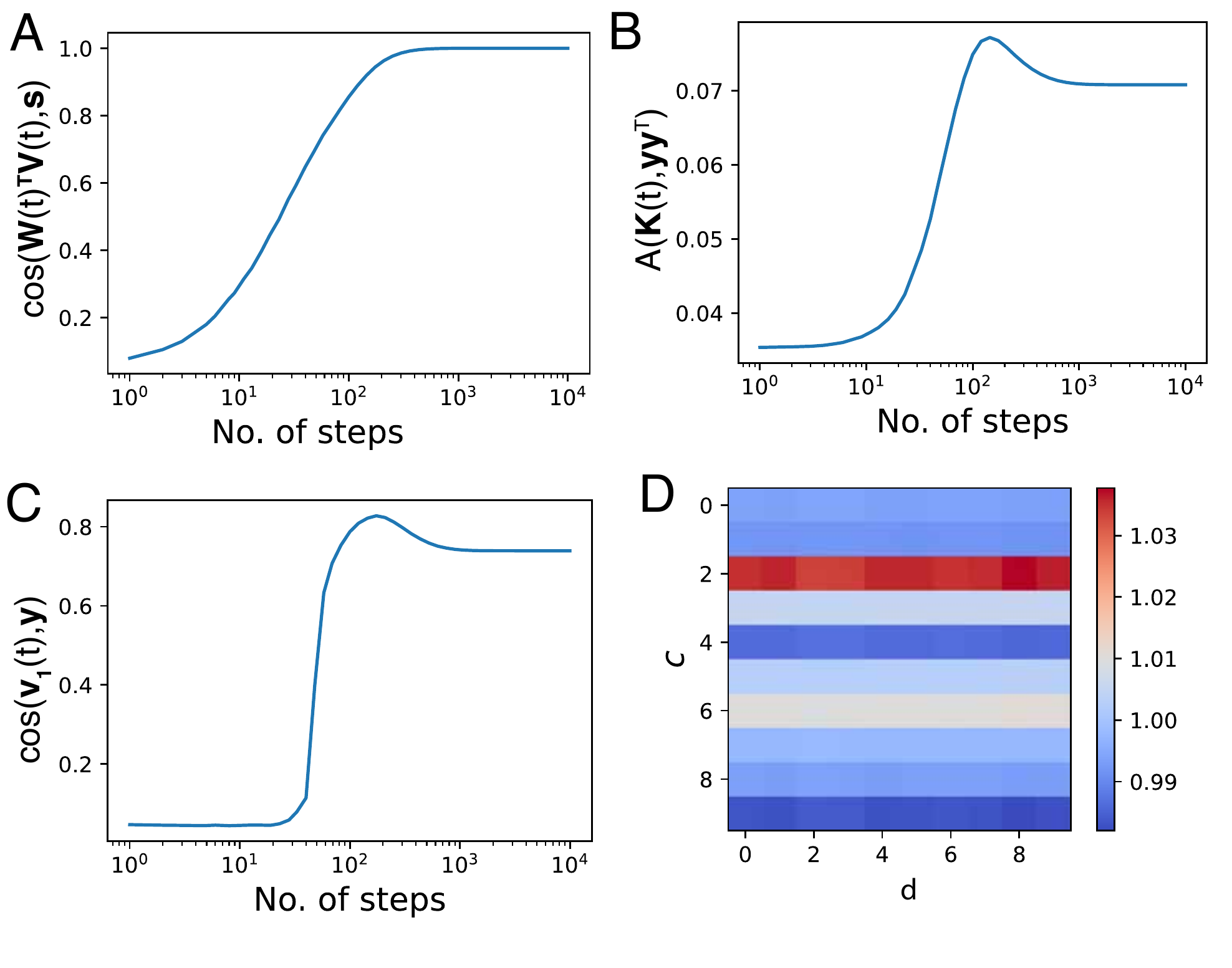}
  \caption{\textbf{Kernel alignment in linear networks during gradient descent.} 
  \\ \textbf{A} Alignment between network weights and teacher weights during learning, as measured by cosine similarity. \textbf{B} NTK becomes more aligned with the task kernel $\bm{yy}^T$. \textbf{C} The top eigenfunction of the NTK becomes more aligned with the target function, as measured by cosine similarity. \textbf{D} The kernel specialization matrix of the linear network shows no specialization. }
  \label{fig:linear}
\end{figure}

We consider deep linear networks of arbitrary depth with $L$ hidden layers and a scalar output, given by $f(\x) = \bm w^{L+1 \top} \bm W^{L} ... \bm W^{1} \bm x$. We assume a small initialization and gradient flow training dynamics. Our results rest on the conservation law identified in prior works in the literature on linear neural networks\cite{advani2020high, cohen_arora, du2018algorithmic}
\begin{align}
    \frac{d}{dt} \left[ \bm W^{\ell} \bm W^{\ell \top} - \bm W^{\ell+1 \top} \bm W^{\ell+1} \right]=0
\end{align}
Under the assumption that the weights are all initialized with small variance this conservation law implies $\bm W^{\ell} \bm W^{\ell \top} \approx \bm W^{\ell+1 \top} \bm W^{\ell+1}$. Starting from the last layer, we infer that $\bm W^{L} \bm W^{L \top} \approx \bm w^{L+1} \bm w^{L+1 \top}$ so that $\bm W^{L}$ is approximately rank-one $\bm W^{\ell} = u(t) \hat{\bm w}^{L+1}(t) \bm r_L(t)$ where $\bm r_L(t)$ and $\hat{\bm w}^{L+1}(t)$ are unit vectors. Repeating this argument inductively generates the conclusion that each layer's weight matrix is rank-one $\bm W^{\ell} = u(t) \bm r_{\ell+1}(t) \bm r_{\ell}(t)^\top$ and thus the kernel has the form
\begin{align}
    K(\x,\x') = u(t)^{2L-2} \x^\top \left[ L \bm r_1(t) \bm r_1(t)^\top +  \bm I  \right] \x'
\end{align}
We provide a derivation in Appendix \ref{app_linear_net_derivation}. The fixed point for $\bm r_1$ is the direction of the linear teacher $\bm \beta$ , since the network must interpolate the data. Thus if one were to evaluate the kernel on the training data, they would obtain 
\begin{equation}
    \K_{\infty} \propto L \bm y \bm y^\top + \bm K_0.
    \label{eq:linear network ntk}
\end{equation}
This is the central result of our theory for linear networks and provides several insights. First, nonlinear activation functions are not necessary for kernel alignment. Second, Even at infinite time, the kernel is not fully aligned with the target function, $\bm{y} \bm{y}^T$, as observed in empirical studies\cite{bahri2021explaining}. Finally, since the first term is linear in network depth, this expression predicts that kernel alignment is more prominent in deeper networks. To test whether this prediction generalizes to nonlinear networks, we trained ReLU two-layer MLPs of different depths on the same MNIST task and found that deeper networks indeed have stronger kernel alignment(Fig.\ref{fig:depth}; details of the experiments are in Appendix.\ref{subsub: two layre relu on mnist}).

\begin{figure}
  \centering
  \includegraphics[scale=0.35]{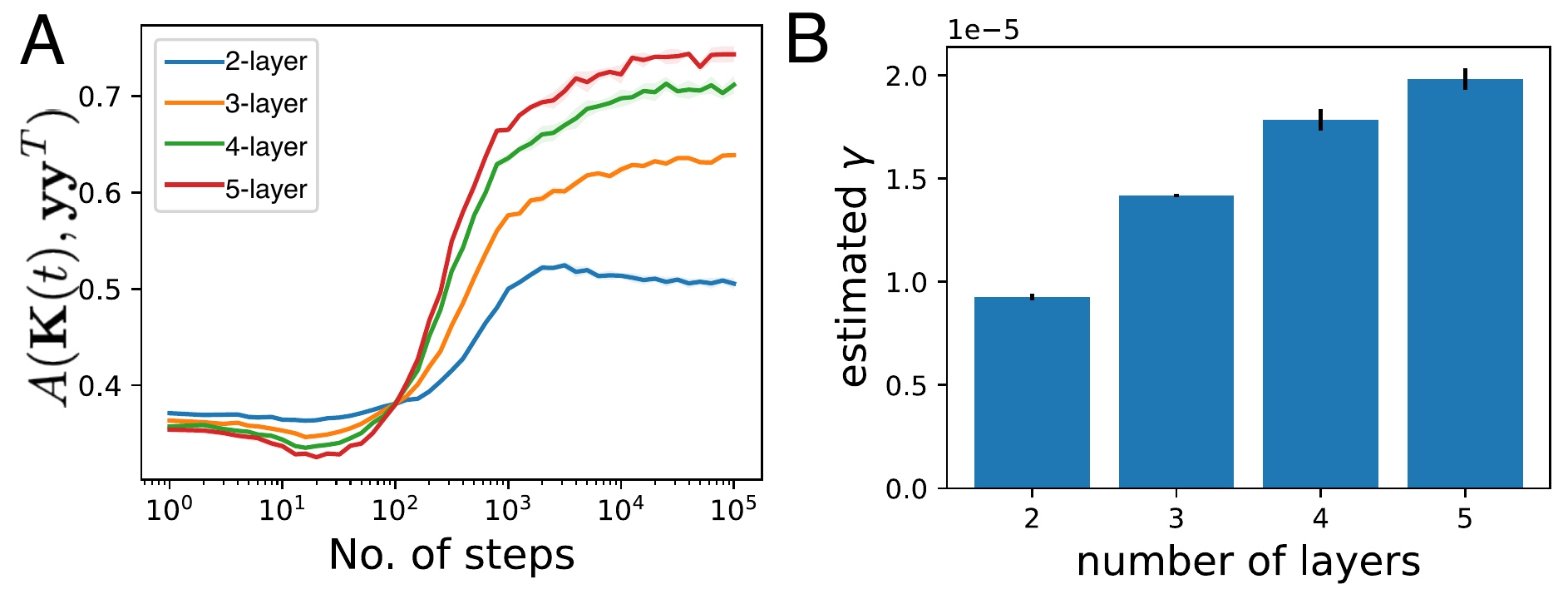}
  \caption{\textbf{Kernel alignment in ReLU MLPs of different depths.} \textbf{A} While all networks start with comparable levels of initial alignment, deeper networks reach higher alignment values at the end of training. \textbf{B} Estimated feature learning rates ($\gamma$) from networks of different depths.}
  \label{fig:depth}
\end{figure}

\subsection{Linear NNs Cannot Develop Specialized Kernels}

We next analyze whether kernel specialization may occur in deep linear networks.  We assume a typical architecture used for multiclass classification, where the network has $L$ shared hidden layers and $C$ linear readouts from the last hidden layer, trained under the setting described in Sec.\ref{subsec:specialization}. In this case and under a weak assumption about symmetry of target functions with respect to $c$, it can be shown that linear networks of arbitrary depth cannot develop specialized kernels(see \ref{section:no specialization in linear}). To test this prediction, we trained a four-layer linear MLP with $10$ output nodes on 10 linear target functions (input vectors were randomly drawn from the unit sphere and the teacher weights for each target function were i.i.d. Gaussian) and computed its KSM. As predicted, the KSM does not show specialization (Fig.\ref{fig:linear}\textbf{D}).

\section{Dynamics of Kernel Alignment in Two-Layer ReLU Networks}
\label{sec:KA in ReLU network}

We next studied how kernel alignment emerges in nonlinear NNs by considering the case of two-layer ReLU networks, a common toy model for studying NN training dynamics in nonlinear networks \cite{ergen2020convex, safran2018spurious, luo2021phase}. In general, expressing the dynamics of the NTK in terms of its $1/\text{width}$ corrections from the static limit requires convoluted schemes \cite{Huang_neural_tangent_hierarchy, dyer2019asymptotics} that are intractable and difficult to interpret. 

We circumvent this issue by exploiting the that fact to study kernel alignment, we are only interested in the structural anisotropy of the kernel in a specific direction (that of the target function). Our approach to track the dynamics of kernel alignment is the following.  We track the bilinear form $B(\bm{z})=\bm{z}^T \bm K \bm{z}$ with $\bm{z} \in S^{P-1}$, a unit vector. Kernel alignment would manifest as $B\left( \bm y / |\bm y| \right)$ growing faster than $\left< B({\bm{z}})\right>_{\bm{z}\sim\text{Unif}(S^{P-1})} \sim \frac{1}{P} \text{Tr}(\bm K)$, where the average is over all unit vectors with a uniform measure for large $P$.

To make kernel alignment dynamics in such nonlinear NNs tractable, we make two heuristic assumptions motivated by empirical observations. First, we assume that the network is sufficiently wide that the sign of preactivations do not change:  $\text{sign}\left(\bm{w_{i}}\cdot\bm{x^{\mu}}\right)$ is static throughout GD dynamics. We also assume that $\forall t>0:$ $\bm{y}^{T}\bm{f}(t)>0$. These assumptions will be justified with simulations.

\subsection{Networks with Scalar Outputs}

For brevity, we write $\phi_i^\mu\equiv \phi(\bm{w_i} \cdot \bm{x^\mu})$. These are the hidden layer features which are dynamic. Then the output of two-layer ReLU network with a scalar output can be written as
\begin{equation} 
    f\left(\bm{x^{\mu}}\right)=\sum_{i=1}^M V_i \phi_i^\mu,
\end{equation}
where $\phi_i^\mu = \text{ReLU}(\bm w_i \cdot\bm x^\mu)$. For simplicity, we consider the task of random binary classification where $\{ \bm{x^\mu} \}$ are drawn i.i.d. from the unit sphere and the target function is a random binary label $y^\mu \in \{-1,1\}$. The effect of more complex data structures is considered below in Sec.\ref{subsection: kernel specialization in relu}. In this setting, we would like to show that kernel alignment occurs and understand it analytically.

A detailed derivation is provided in Appendix \ref{appd: KA in two layer relu} and we provide a sketch here. For this network, $\bm K$ has two components, given by
\begin{align}
    K_{\mu,\nu} &= \nabla_{\bm{V}} f^\mu \cdot \nabla_{\bm{V}} f^\nu + \nabla_{\bm{W}} f^\mu \cdot \nabla_{\bm{W}} f^\nu \\
    & \equiv (\bm K_{\bm{V}})_{\mu,\nu} + (\bm K_{\bm{W}})_{\mu,\nu},
\end{align}
The first component is contributed by  $\nabla_{\bm V} f$ and the second by $\nabla_{\bm W} f$. Denote their respective contribution to $B(\bm z)$ as $\alpha(\bm{z}) = \bm{z}^T \bm{K_V} \bm{z}$ and $\beta(\bm{z}) =  \bm{z}^T \bm{K_W} \bm{z}$. We show in Appendix \ref{app:relu_align_theory} that gradient descent generates the following dynamics
\begin{align}
    \frac{d\alpha}{dt}\left(\bm y / |\bm y| \right) &\approx \frac{1}{P}\eta\left[\left(\bm{y}-\bm{f}\right)^{T}\bm{y}\right]\left[\bm{y}^{T}\bm{f}\right] \nonumber \\
    \left\langle \frac{d\alpha}{dt}\left(\bm{z}\right)\right\rangle _{\bm{z}} & \approx \frac{1}{P}2\eta\left(\bm{y}-\bm{f}\right)^{T}\bm{f} \label{eq:two layer dynamics} \\
    \forall \bm{z}: \frac{d\beta}{dt}\left(\bm{z}\right) & \approx  \frac{1}{2}\eta(\bm{y}-\bm{f})^{T}\bm{f} \nonumber
\end{align}

While these equations are not closed (they depend on $\bm{f}(t)$), it is clear that the dynamics of $\beta(\bm{z})$, and by extension $\bm{K_W}$, are independent of $\bm{z}$ and therefore isotropic. On the other hand, we can consider the anisotropy of $\alpha$ by considering early stages of learning where $\bm{y}-\bm{f}\approx \bm{y}$, yielding $\frac{d\alpha}{dt}\left( \bm y/|\bm y|\right) \approx \eta \bm{y}^{T}\bm{f}$ and $\left\langle \frac{d\alpha}{dt}\left(\bm{z}\right)\right\rangle _{\bm{z}} \approx \frac{1}{P}2\eta\bm{y}^{T}\bm{f}$. Note that $|\bm y|^2 = P$ since $y_\mu \in \{ \pm 1\}$. These results indicate that $\bm {K_V}$ grows $O(P)$ times faster in the direction of $\bm{y}$ than in the other directions. The NTK, which is the sum of $\bm{K_V},\bm{K_W}$, therefore develops an anisotropy in the form of kernel alignment with $\bm{y}$.

\begin{figure}[h]
    \centering
    \includegraphics[width=\linewidth]{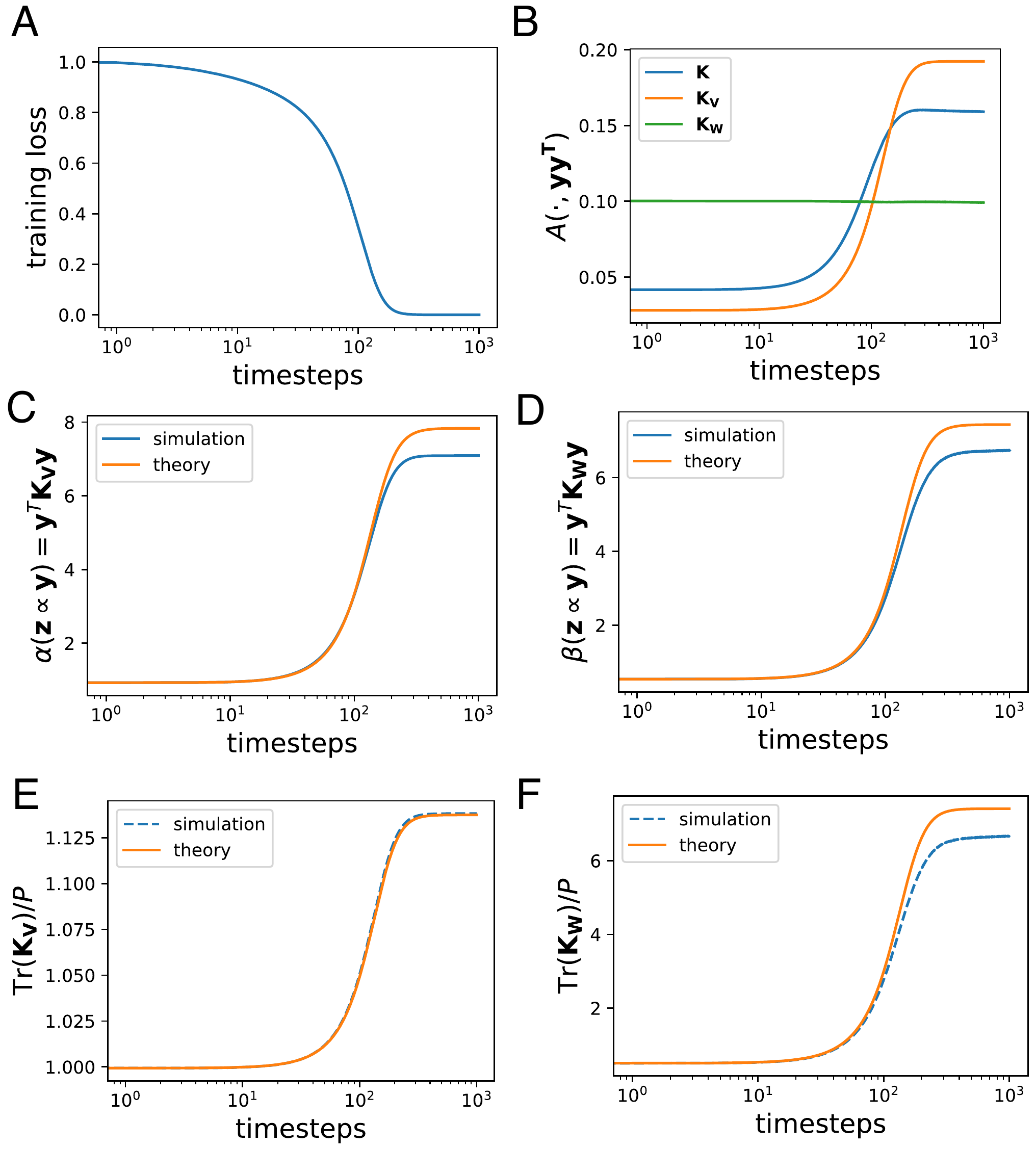}
    \caption{\textbf{Kernel alignment in two-layer ReLU network is well characterized by theory.} \\
    \textbf{A} Training loss over time. \textbf{B} Alignment between the NTK ($\bm{K_V} + \bm{K_W}$) and the target functions increases before saturating. As predicted by the theory, alignment is driven by anisotropy in $\bm{K_V}$ while changes to $\bm{K_W}$ are isotropic. \textbf{C} Dynamics of the projection of $\bm{K_V}$ on the target functions. \textbf{D} Dynamics of the projection of $\bm{K_W}$ on the target functions. \textbf{E} Dynamics of the trace of $\bm{K_V}$ (same as $\langle \bm{z}^T \bm{K_V} \bm{z} \rangle_{\bm z}$. Notice that the growth is much weaker than the growth in the direction in $\bm{y}$ (panel C). \textbf{F} Dynamics of the trace of $\bm{K_W}$. Notice that the growth is of the same magnitude than the growth in the direction in $\bm{y}$ (panel D).}
    \label{fig:two layer relu}
\end{figure}

To quantitatively test our theory, we trained (see details in Appendix \ref{subsub: two layer relu random binary details})  two-layer ReLU networks on the binary classification task using gradient descent and found that Eq.\ref{eq:two layer dynamics} nicely capture the dynamics and overall strength of kernel alignment in the networks(Fig.\ref{fig:two layer relu}). First, as predicted, kernel alignment is driven by alignment in the $\bm{K_V}$ component, as predicted(Fig.\ref{fig:two layer relu}\textbf{B}). In addition, $\bm{K_V}$ grows at a much faster rate in the direction of $\bm{y}$(Fig.\ref{fig:two layer relu}\textbf{C}) than in other directions(Fig.\ref{fig:two layer relu}\textbf{E}). On the other hand, $\bm{K_W}$ grows at the same rate in all directions(Fig.\ref{fig:two layer relu}\textbf{D,F}).

\subsection{Dynamics of Kernel Specialization in a Classification Setting}
\label{subsection: kernel specialization in relu}
To analytically understand kernel specialization, we now extend the above analysis to NNs with multiple output heads. For concreteness, we consider a two-layer ReLU network with $C$ output heads. Output from the $c$th head is given by
\begin{equation}
    f_c^\mu = f_c(\bm{x}^\mu)=\sum_{i=1}^M V_i^c \phi_i^\mu.
\end{equation}
For the task, we assume the scenario of training the network to separate a mixture of $C$ Gaussian distributions, $\{\mathcal{N}(\bm{\mu_c}, \sigma^2 \mathbf{I})\}_{c=1,...,C}$. From each distribution, $P/C$ samples are drawn into the training set. We assume $\{\bm{\mu}_c\}$ to be unit vectors that are pairwise perpendicular. Each head has a separate target function, $\bm{y}_c$. We consider the case where the target functions encode the correct class of each input. I.e. $y^\mu_c=1$ if $\bm{x}^\mu$ is in class $c$ and $0$ otherwise. Here we present the simplistic but illuminating limit where $\sigma^2 \rightarrow 0$ and $\bm{x}^{\mu \in \text{class c}}=\bm {\mu_c}$ (the case of finite but small $\sigma^2$ is discussed in Appendix \ref{apd:specialization}, but the derivation has the same flavor as the one here). At this limit, the input correlation structure is simply $\bm{x}^\mu \cdot \bm{x}^\nu=1$ if they are in the same class and zero otherwise (this is possible for $C<N$).

Importantly, this structure in input space induces an analogous structure in $\{ \bm{D}^\mu\}_{\mu=1,...,P}$, where $D^\mu_i=d_x\phi(x)|_{x=\bm{w_i}\cdot \bm{x^\mu}}$. We note that the covariance of $D^{\mu}, D^{\nu}$ over random hidden weight vectors $\bm w$ gives 
\begin{align}
    \text{Cov}(D^{\mu} , D^{\nu}) \approx \delta_{ y_\mu , y_{\nu}} \frac{3}{4}.
\end{align}
Crucially, as we proceed to show, \textit{this class-dependent structure in $D^\mu_i$ is necessary for kernel specialization in this setting}. This is also consistent with our finding that linear networks cannot specialize since $D^\mu_i=1$ for all $i,\mu$ for linear nets.

To detect kernel specialization, we consider the structure
of the $c$-th subkernel by tracking the bilinear form $B^c(\bm y_d)=\bm y_d^T \bm{K}^{c,c} \bm y_d$. In order to show kernel specialization,  $B^c(\bm y_d)$ must grow faster for $d=c$ then $d \neq c$. We first observe that the subkernel has two components, similar to the scalar-output network case, given by $K^{c,c}_{\mu,\nu} = \left(K^{c,c}_{\bm{V}}\right)_{\mu,\nu} + \left(K^{c,c}_{\bm{W}}\right)_{\mu,\nu}$. Note that $\bm K^{c,c}_{\bm{V}}$ is the same for all $c$ and therefore cannot contribute to kernel specialization.  By exclusion, kernel specialization must arise from $\bm K^{c,c}_{\bm{W}}$. We can thus simply our analysis of anisotropic dynamics of $\bm{K^{c,c}}$ by by focusing on its component coming from $\bm K^{c,c}_{\bm{W}}$, given by $B^c_{\bm{W}}(\bm{y}_d) \equiv \bm{y}_d^T \bm K^{c,c}_{\bm{W}} \bm{y}_d$. At the $\sigma^2 \rightarrow 0$ limit, points from each class trivially collapse to a single point, allowing us to write $\forall \mu \in \text{class c}:D^\mu_i=D^c_i,\phi^\mu_i=\phi^c_i$. One can show that the dynamics of it follow
\begin{align}
    \eta^{-1}\frac{d B^c_{\bm{W}}(\bm{y}_d)}{dt} &=  \sum_{i=1}^M \frac{d B^c_{\bm{W}}(\bm{y}_d)}{d (\bm V_c)_i} \frac{d (\bm V_c)_i}{dt}\nonumber \\
    %=&\sum_i \sum_{\mu \in \text{class d}} D^\mu_i ({\bm{V_c}})_i \sum_{\nu  \in \text{class c}} \phi^\nu_i \\
    =& \left( \frac{P}{C} \right)^2 \sum_{i=1}^M D^{d}_i ({\bm{V_c}})_i \phi^{c}_i.
    \label{eq:dynamics of B^c_W}
\end{align}
The sum over neuron indices $i$ highlights how the correlation structure between $\{\bm D^c_i \}_{i=1,...,M}$ and $\{\bm D^d_i \}_{i=1,...,M}$ lead to kernel specialization. If $d=c$, then $B_c(\bm y_d)$ increases as 
\begin{equation}
    \eta^{-1}\frac{d B^c_{\bm{W}}(\bm{y}_d)}{dt} = \left( \frac{P}{C} \right)^2 f_c(\bm{x}^{\mu \in \text{class c}}) >0.
\end{equation}
It follows from our assumption of $\sum_\mu y^\mu_c f(\bm{x}^\mu) > 0$ that $f_c(\bm{x}^{\mu \in \text{class c}})>0$ (since all data points from each class collapse to one point, this is the same regardless of $\mu$).  %In our case, the input correlation structure  makes them weakly coupled (see Appendix.\ref{eq:bernoulli ansatz}). 
Now we consider $d \neq c$. The sum in equation \ref{eq:dynamics of B^c_W} will always contain fewer terms than the full sum over $i \in \{1,...,M \}$ since $D_{i}^{\mu} \in \{0,1\}$. We approximate $D_{i}^\mu$ as independent of $\phi_i^\nu$, giving approximately half of the terms $\{ (\bm{V}_c)_i \phi^c_i \}_{i=1,...,M}$ (see further discussion in Appendix.\ref{eq:bernoulli ansatz}). This motivates the following approximation
\begin{equation}
    \frac{d B^c_{\bm{W}}(\bm{y}_d)}{dt} \approx \frac{1}{2}  \frac{d B^c_{\bm W}(\bm y_c)}{dt}.
    \label{eq:nonspecialized alignment}
\end{equation}
\begin{figure}[!ht]
  \centering
  \includegraphics[width=\linewidth]{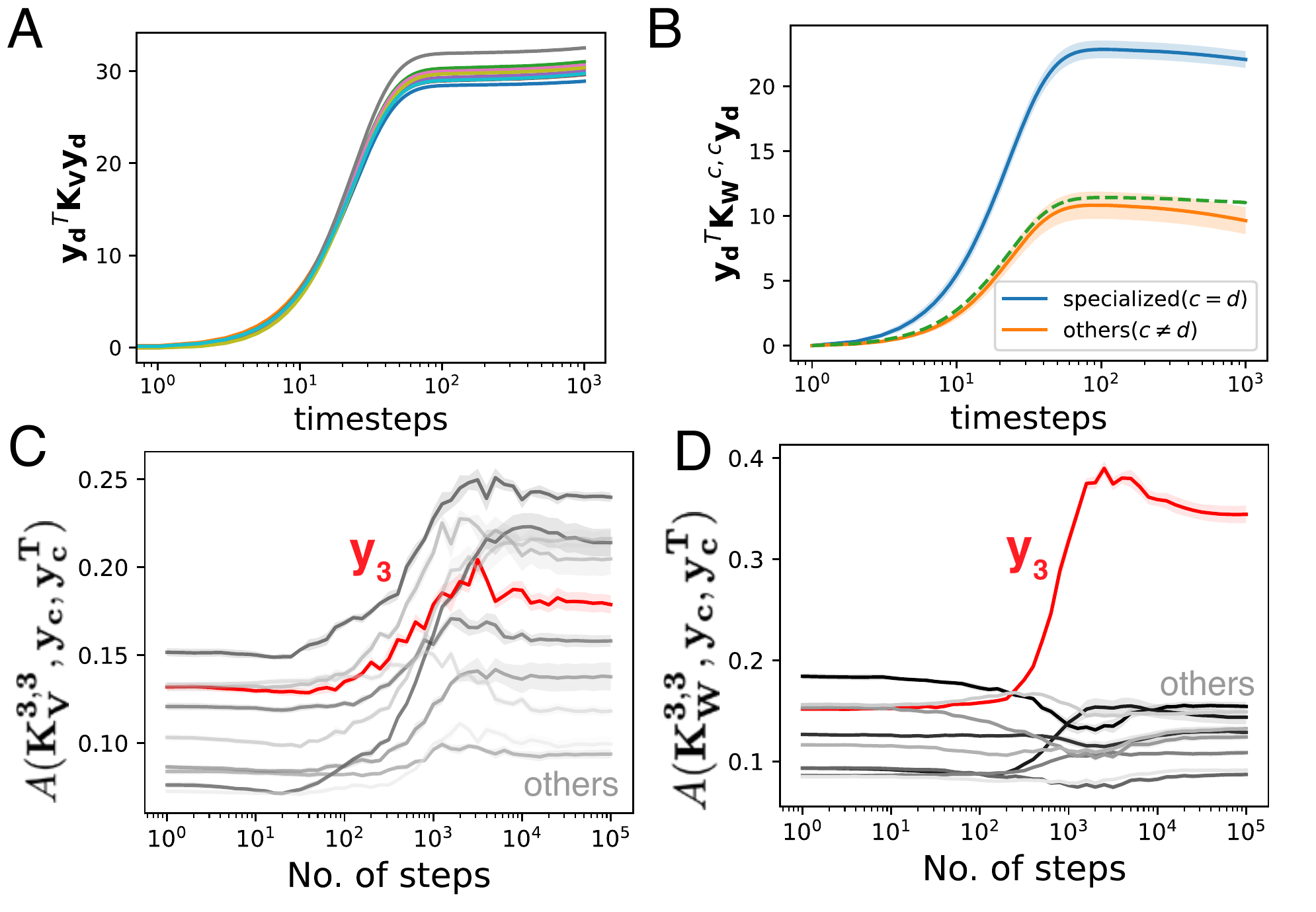}
  \caption{\textbf{Total kernel alignment is driven by both indiscriminate alignment and kernel specialization.} \\
  \textbf{A}. In a ReLU network trained to classify a mixture of 10 Gaussians, the $\bm{K_V}$ component of the NTK grows in the direction of all 10 target functions at approximately the same rate. This component is the same for all subkernels. Each trace is the projection on a different target function, with its initial value subtracted. \textbf{B}. In the same setting as \textbf{A}, the $\bm{K_W}^{c,c}$ component of the NTK grows in the direction of its corresponding target function (``specialized") at around twice the rate of that in the direction of other target functions (``others"), as predicted by our theory. Green dashed line is half the specialized rate. In all cases the initial value is subtracted from the entire trace. The error bars are standard deviations over different $c,d$. \textbf{C} In a ReLU network trained on the MNIST task, $\bm{K_V}$ does not specialize as its alignment with all target functions grows at a similar rate. \textbf{D}. In the same setting as \textbf{C}, the $\bm{K_W}^{c,c}$ component of the NTK becomes highly specialized.\\
  Average of 5 runs; standard error shown as color contour.}
  \label{fig:specialization}
\end{figure}
In other words, $K^{c,c}_{\bm{W}}$ grows approximately twice as fast in the direction of $\bm{y}_c$ compared to $\bm{y}_d$, indicating kernel specialization. Since, $K^{c,c}_{\bm{V}_c}$ does not differentiate between different target functions, the full subkernel $K^{c,c}_{\bm{W}} + K^{c,c}_{\bm{V}}$ becomes specialized over time. To test our theory quantitatively, we trained two-layer ReLU networks with 10 output heads to classify a mixture of 10 Gaussians (with $\sigma^2=0.01$, see details in Appendix\ref{subsub: two layer relu mixture}) and measured how the two components of each subkernel ($\bm{K^{c,c}}$), $\bm{K_{V}^{c,c}}$ and $\bm{K_W^{c,c}}$ grow in the directions of the 10 target functions. As predicted by our theory, the $\bm{K_{V}^{c,c}}$ component grows at the same rate in the direction of all target functions(Fig.\ref{fig:specialization} \textbf{A}) and therefore does not contribute to kernel specialization (it still contributes to alignment). On the other hand, the $\bm{K_W^{c,c}}$ component grows at approximately twice the rate in the direction of its corresponding target function ($\bm{y_c}$) compared to in the directions of other target functions ($\bm{y_c\neq d}$, Fig.\ref{fig:specialization} \textbf{B})), as predicted by Eq.\ref{eq:nonspecialized alignment}.

We also verified these qualitative predictions in more practical settings by training a two-layer ReLU MLP on the 10-class MNIST classification task and plotting how well $K^{c,c}_{\bm{W}}, K^{c,c}_{\bm{V}}$ align to the 10 target functions, respectively (Fig.\ref{fig:specialization}, $c=3$ in the figures). As predicted, $K^{3,3}_{\bm{V}}$ becomes aligned with 10 target functions to a similar extent (Fig.\ref{fig:specialization}\textbf{C}) while $K^{3,3}_{\bm{W}}$ preferentially aligns with $\bm{y}^3$(Fig.\ref{fig:specialization}\textbf{D}). 

%We can now consider how input correlation between points of the same class affects the strength of specialization. This correlation does not affect the rate of kernel growth in the direction of $\bm y_c$. If $\bm x$ from different classes are anticorrelated, $\rho_{\mu,\nu}$ will be negative for $\bm x^\mu, \bm x^\nu$ from different classes. The sum in Eq.\ref{eq:dynamics of B^c_W} would be close to picking out all the zero elements from $\{ (\bm{V}_c)_i \phi^{\nu \in \text{class c}}_i \}_{i=1,...,M}$. Thus the growth rate in the direction of $\bm{y}_{d \neq c}$ is close to zero, and specialization is more stronger. Vice cersa, if $\bm x$ from different classes are correlated, it follows that $K^{c,c}_{\bm{W}}$ grows in a similar rate for all ${\bm y_c}_{c=1,...,C}$ and shows weak specialization.

%%%%%%%%%%%%%%%%%%%%%%%%%%%%%%%%%%%%%%%%%%%%%%%%%%%%%%%%%%%%%%%%%%%%%%%%
\section{Conclusions}

This work demonstrated how kernel alignment emerges during NN learning dynamics and in turn accelerates learning, both through experiments and theory. Kernel alignment induces an anisotropic structure in the NTK over time. We first demonstrated empirically that learning is accelerated by this anisotropy in a way that cannot be simply accounted for by an increase in the scale of the kernel. We analytically studied the optimal feature evolution model to show that aligning the NTK with the target function is beneficial. For the first time to our best knowledge, we demonstrated empirically that in NNs with multiclass outputs, alignment manifests in the form of kernel specialization, where the subkernel corresponding to each output head aligns to its corresponding target function, but not others.

We then developed theoretical analyses of how the NTK acquires an anisotropic structure that aligns with the target function during NN training. By analyzing deep linear networks and two-layer ReLU networks, we give tractable analytical descriptions of the alignment dynamics of the NTK. Our analyses suggested that alignment occurs faster and more strongly in deeper networks and that specialization is a phenomenon that requires nonlinear activation functions and specific structures in the data.

\subsection*{Limitations and Future Directions}
A primary limitation of our work is the set of simplifying assumptions that we have taken: focusing on the mean-squared error loss and assuming the simple data structures for the theory in ReLU networks. Furthermore, our derivations rely on heuristic assumptions and ansatzes, which should be made rigorous. This may be possible through performing quenched averages over initial weights and data points using techniques from statistical physics.

Another limitation is that we have focused on the kernel evaluated on the training set through the $P\times P$ Gram matrix. It will be interesting to extend some of the analysis to the test set to evaluate how kernel alignment affects generalization.

\bibliography{example_paper}
\bibliographystyle{icml2022}

%%%%%%%%%%%%%%%%%%%%%%%%%%%%%%%%%%%%%%%%%%%%%%%%%%%%%%%%%%%%%%%%%%%%%%%%%%%%%%%
%%%%%%%%%%%%%%%%%%%%%%%%%%%%%%%%%%%%%%%%%%%%%%%%%%%%%%%%%%%%%%%%%%%%%%%%%%%%%%%
% APPENDIX
%%%%%%%%%%%%%%%%%%%%%%%%%%%%%%%%%%%%%%%%%%%%%%%%%%%%%%%%%%%%%%%%%%%%%%%%%%%%%%%
%%%%%%%%%%%%%%%%%%%%%%%%%%%%%%%%%%%%%%%%%%%%%%%%%%%%%%%%%%%%%%%%%%%%%%%%%%%%%%%
\newpage
\appendix
\onecolumn
\section{Derivation of NTK for Linear Network with Small Initialization}\label{app_linear_net_derivation}

The neural tangent kernel for the deep linear network $f(\x) = \bm w^{L+1 \top} \bm W^{L} ... \bm W^{1} \x$ is 
\begin{align}
    K(\x,\x') = \sum_{\ell=1}^{L+1} \left< \frac{\partial f(\x)}{\partial \bm W^{\ell}}, \frac{\partial f(\x')}{\partial \bm W^{\ell}} \right>_{F} 
\end{align}
Using the fact that, from small initialization, $\bm W^{\ell} = u(t) \bm r_{\ell+1}(t) \bm r_{\ell}(t)^\top$, we find for $\ell > 1$
\begin{align}
    \frac{\partial f(\x)}{\partial \bm W^{\ell}} &= \left( \prod_{\ell' > \ell} \bm W^{\ell'}\right)^\top \x^\top \left( \prod_{\ell' < \ell} \bm W^{\ell'} \right)^\top = u(t)^{L-1} \x^\top \bm r_1(t)  \bm r_{\ell+1}(t)  \bm r_{\ell}(t)^\top \nonumber
    \\
    \left< \frac{\partial f(\x)}{\partial \bm W^{\ell}}, \frac{\partial f(\x')}{\partial \bm W^{\ell}} \right> &= u(t)^{2L-2} \x^\top \bm r_1(t) \bm r_1(t)^\top \bm x' \left<  \bm r_{\ell+1}(t)  \bm r_{\ell}(t)^\top, \bm r_{\ell+1}(t) \bm r_{\ell}(t)^\top \right>_F \nonumber
    \\
    &= u(t)^{2L-2} \x^\top \bm r_1(t) \bm r_1(t)^\top \bm x'
\end{align}
For $\ell = 1$ we have
\begin{align}
    \frac{\partial f(\x)}{\partial \bm W^{1}} = \left( \prod_{\ell' > 1} \bm W^{\ell'} \right)^{\top} \bm x^\top \implies \left< \frac{\partial f(\x)}{\partial \bm W^1}, \frac{\partial f(\x')}{\partial \bm W^1}   \right> = u(t)^{2L-2} \x \cdot \x'   
\end{align}
Thus, adding the contributions from each of the $L+1$ layers, we find
\begin{align}
    K(\x,\x') = u(t)^{2L-2}  \x^\top \left[ L \bm r_1(t) \bm r_1(t) + \bm I \right] \x'
\end{align}
This shows that the relative size of the rank-one spike will be controlled by the depth of the network, $L$. 

\section{Optimal Feature Evolution Induces Kernel Alignment}\label{app_optimal_evolution}

Let $\bm\Delta = \bm f - \bm y$ and let $\bm\Psi \in \mathbb{R}^{N \times P}$ represent the feature matrix whose inner product gives the kernel $\K = \bm\Psi^\top \bm\Psi$. We will first discuss a discrete time dynamical system before taking a gradient flow limit. Thus, we index $\bm\Delta_t$ as the error at time $t$ and $\bm\Psi_t$ as the features at time $t$. We consider the following simulataneous updates to $\bm\Psi_t$ and $\bm\Delta_t$

\begin{align}
    &\bm\Delta_{t+1}(\bm\Psi_t) = \bm\Delta_t - \eta \bm\Psi_t^\top \bm\Psi_t \bm\Delta \nonumber
    \\
    \bm\Psi_{t+1} &= \bm\Psi_t - \eta \gamma \frac{\partial}{\partial \bm\Psi_t} || \bm\Delta_{t+1}(\bm\Psi_t) ||^2\\
    &= \bm\Psi_t - \frac{1}{2} \eta \gamma \frac{\partial}{\partial \bm\Psi_t} || \bm\Delta_t - \eta \bm\Psi_t^\top \bm\Psi_t \bm\Delta_t ||^2
\end{align}

Expanding the last term and computing the derivative gives

\begin{align}
    &\frac{1}{2}\frac{\partial }{\partial \bm\Psi_t} || \bm\Delta_t - \eta \bm\Psi_t^\top \bm\Psi_t \bm\Delta_t ||^2 \\
    &= \frac{1}{2} \frac{\partial }{\partial \bm\Psi_t} \left[ ||\bm\Delta_t||^2 - 2 \eta \bm\Delta_t^\top \bm\Psi_t^\top \bm\Psi_t \bm\Delta_t + \eta^2 \bm\Delta^\top \bm\Psi_t^\top  \bm\Psi_t\bm\Psi_t^\top \bm\Psi_t \bm\Delta \right] \nonumber
    \\
    &= - \eta \bm\Psi_t \bm\Delta_t \bm\Delta_t^\top + \eta^2 \bm\Psi_t \bm\Psi_t^\top \bm\Psi_t \bm\Delta_t \bm\Delta_t^\top + \eta^2 \bm\Psi_t \bm\Delta_t \bm\Delta_t^\top \bm\Psi_t^\top \bm\Psi_t \nonumber
    \\
    &= - \eta \bm\Psi_t \bm\Delta_t \bm\Delta_t^\top + O(\eta^2)
\end{align}

Taking the $\eta \to 0$ limit while taking the distance in time between adjacent steps to zero, we find the following gradient flow dynamics
\begin{align}
    \dot{\bm\Delta}(t) = - \eta \bm\Psi(t)^\top  \bm\Psi(t) \bm\Delta(t) \ \quad \ \dot{\bm\Psi}(t) = \gamma \eta \bm\Psi(t) \bm\Delta(t) \bm\Delta(t)^\top 
\end{align}

This is a collection of coupled nonlinear ODEs. The $\gamma \to 0$ limit recovers lazy learning where the features do not evolve. Increasing $\gamma$ increases the rate at which features evolve, thus we name it the \textit{feature learning rate}. Despite the nonlinearity, we will attempt to solve these equations to gain insight into such optimal feature updates. The key trick is that the equations can be decoupled through the use of a conservation law. To motivate the conservation law, consider the scalar version of these differential equations
\begin{align}
    \dot \Delta = - \psi^2 \Delta  \quad
    \dot \psi = \gamma \psi \Delta^2
\end{align}

Note that $\frac{1}{2} \frac{d}{dt} \Delta^2 = - \Delta^2 \psi^2$ while $\frac{1}{2} \frac{d}{dt} \psi^2 = \gamma \psi^2 \Delta^2$. A particular linear combination of these terms reveals a conservation law 
\begin{equation}
    \frac{\gamma}{2} \frac{d}{dt} \Delta^2 + \frac{1}{2} \psi^2 = \frac{1}{2} \frac{d}{dt}\left[ \gamma \Delta^2 + \psi^2  \right] =  - \gamma \Delta^2 \psi^2 + \gamma \Delta^2 \psi^2 = 0
\end{equation}

Thus $\gamma \Delta^2 + \psi^2$ is a conserved quantity throughout the dynamics. This indicates that, in $(\Delta, \psi)$ space, the trajectory can only move along an ellipse, with the ratio of axis lengths determined by $\sqrt{ \gamma }$. Using the conservation law, we can introduce a constant $C = \gamma \Delta^2 + \psi^2 = \gamma \Delta_0^2 + \psi_0^2$, where $\Delta_0$ and $\psi_0$ are the initial values. We note the similarity between these elliptical differential equations and the \textit{hyperbolic} geometry of gradient descent in deep linear neural networks \cite{Saxe_McClelland_Ganguli}, where the conservation laws have the form $a^2 - b^2 = C$. Using our discovered elliptical conservation law, the differential equations can now be decoupled
\begin{equation}
    \dot\Delta = - (C - \gamma\Delta^2) \Delta \quad \dot \psi =  \psi ( C  - \psi^2)
\end{equation}

Letting $u = \frac{1}{2} \Delta^2$ and $v = \frac{1}{2} \gamma^2$, we find $\dot u = -u(C-\gamma u)$ and $\dot v= v(C - v)$, which give solutions of the form
\begin{equation}
    u = \frac{CA}{A + e^{2Ct}} \quad v = \frac{C B}{B + e^{-2Ct}}
\end{equation}

for constants $A$ and $B$ determined by the initial conditions. This indicates that the loss and the kernel power increases as logistic functions. Since $u$ represents the loss, this indicates that at large times, a scaling of $u \sim \exp( - 2 (\psi_0^2 + \gamma \Delta_0^2) t)$ is obtained, which improves with increasing $\gamma$. 

The scalar case was illuminating since it allowed us to identify a conservation law and solve the differential equation. We now aim to extend this argument to arbitrary dimensional matrices $\bm\Psi(t) \in \mathbb{R}^{N \times P} $ and vectors $\bm\Delta(t) \in \mathbb{R}^P$. Inspired by the elliptical geometry in the scalar case, we make the following ansatz that $\bm C= \gamma \bm\Delta \bm\Delta^\top + \bm\Psi^\top \bm\Psi$ is conserved. Indeed, explicit differentiation reveals this to be the case.

\begin{align}
    \frac{d}{dt} &\left[ \gamma \bm\Delta \bm\Delta^\top + \bm\Psi^\top \bm\Psi \right] \nonumber
    \\
    &= - \eta \gamma \bm\Psi^\top \bm\Psi\bm\Delta \bm\Delta^\top - \eta \gamma \bm\Delta \bm\Delta^\top \bm\Psi^\top \bm\Psi + \eta \gamma \bm\Delta\bm\Delta^\top \bm\Psi^\top \bm\Psi + \eta \gamma \bm\Psi^\top \bm\Psi \bm\Delta \bm\Delta^\top = \bm 0 
\end{align}

Thus $\bm C = \gamma \bm\Delta \bm\Delta^\top + \bm\Psi^\top \bm\Psi$ is a conserved matrix. We can use this fact to again decouple the dynamics
\begin{equation}
    \dot{\bm \Delta}  = - \eta ( \bm C - \gamma \bm\Delta \bm\Delta^\top ) \bm\Delta  \quad \dot{\bm\Psi} = \eta \bm\Psi ( \bm C - \bm\Psi^\top \bm\Psi) 
\end{equation}

Positive $\gamma$ has the effect of accelerating convergence of $\bm\Delta$ to zero, while the initial condition and final conditions can be explicitly related $\bm C = \K_{\infty} =  \gamma \bm y \bm y^\top + \bm\Psi_0^\top \bm\Psi_0$,  demonstrating that the kernel will align more with the labels after training. The dynamics of the loss and the kernel can be examined in the eigenbasis of $\bm C$. Let $\bm C = \K_{\infty} = \sum_k c_k \bm v_k \bm v_k^\top$ and let $\bm\Delta = \sum_k \delta_k \bm v_k$ and $\K = \bm\Psi^\top \bm\Psi = \sum_{k,\ell} A_{k,\ell} \bm v_k \bm v_{\ell}^\top$ for symmetric matrix $\bm A$. This generates the following differential equations
\begin{align}
\frac{d}{dt} \ln \delta_k &= - \eta c_k + \eta \gamma \sum_{\ell} \delta_\ell^2 \nonumber
\\
\dot A_{k,\ell} &= \eta  A_{k,\ell} (c_k + c_\ell) - 2 \eta\sum_j A_{k,j} A_{j,\ell} 
\end{align}

To zero-th order in $\gamma$, the loss scales like $L_t = \sum_k \left( \bm v_k^\top \bm y \right)^2 \exp( - 2 \eta c_k t)$, which in general will  decay more quickly than the loss for the frozen kernel, since $\bm K^{\infty} $ is more aligned with $\bm y$ than $\bm K_0$. When $\gamma$ is small but non-negligible, we expect $(\bm v_1^\top \bm y)^2  \gg \left( \bm v_k^\top \bm y \right)^2$ for $k>1$. We thus get a loss that looks like $L_t = D e^{-2\eta c_1 t} + \sum_{k > 1} \left( \bm v_k^\top \bm y \right)^2 \exp( - 2 \eta c_k t)$. The first term dominates at small times since it has a large prefactor constant $D = (\bm v_1^\top \bm y)^2$, however once $t \approx 1/c_1$, the tail sum dominates and the loss falls as a power law, with a possibly improved exponent due to better alignment. 

Now, let's consider the kernel's dynamics. First, at small times $A_{k,\ell}$ is non-diagonal since $\K_0 $ and $\K_{\infty}$ do not necessarily commute. These off diagonal terms will eventually decay due to the $- \sum_j A_{k,j} A_{j,\ell}$ term. Once $A$ is approximately diagonal, the dynamics for the diagonal terms are
\begin{equation}
    \dot A_{k,k} =2 \eta A_{k,k} c_k - 2 \eta A_{k,k}^2 
\end{equation}

This is identical to the scalar equations studied above which we can solve exactly
\begin{equation}
    A_{k,k}(t) = \frac{B_k c_k}{B_k + e^{-2 \eta c_k t}} 
\end{equation}
for some constants $B_k$ determined by the initial values $A_{kk}(0)$. Thus, $A_{k,k}(t)$ increase as logistic functions with a time constant given by $c_k$. Thus, the kernel is approximately
\begin{equation}
    \K(t) \sim \sum_{k} A_{kk}(t) \bm v_k \bm v_k^\top
\end{equation}
which gives an alignment of 
\begin{equation}
    \left< \bm y \bm y^\top , \K(t) \right> =\frac{1}{\gamma} \left<  \K_{\infty} - \K_0 , \K(t) \right>_F = \frac{1}{\gamma} \sum_k \left( c_k - A_{k,k}(0) \right)  A_{kk}(t) 
\end{equation}
which increases as a weighted sum of logistic functions. The norm of the kernel grows as $||\K(t)||_F^2 = \sum_k A_{k,k}(t)^2$ so the alignment curve has the form
\begin{equation}
    A(t) = \frac{\sum_k \left( c_k - A_{k,k}(0) \right)  A_{kk}(t) }{\sqrt{\sum_k A_{k,k}(t)^2} \sqrt{||\K_0||^2 - 2 \sum_k c_k A_{k,k}(t) + \sum_k A_{k,k}(t)^2 }} .
\end{equation}

% \subsection{Two-layer linear networks with scalar output shows kernel alignment}
% \label{section:two layer linear network}
% We numerically tested the approximation taken in Sec.\ref{sec:alignment in deep linear networks} by comparing its eigenspectrum and task spectrum with those of the kernel. 

% \begin{figure}
%     \centering
%     \includegraphics[scale=0.4]{figures/linear_kernel_approx.pdf}
%     \caption{Comparing the eigenspectra and task spectra of the NTK for a two-layer linear network to test the theory in section \ref{sec:alignment in deep linear networks}, showing good consistency. \textbf{A}. Eigenspectra. \textbf{B}. Task spectra.}
%     \label{fig:linear kernel approx}
% \end{figure}

\section{Rescaling Alters Feature Learning Rate}\label{app:rescaling}

In the paper, we consider the following rescaling of the output function $g(\x) = \frac{1}{\gamma} f(\x)$ and let the learning rate be $\eta = \eta_0 \gamma^2$. With this choice, gradient flow on the loss $L = \sum_\mu \ell( \frac{1}{\gamma} f(\x^\mu) ,y^\mu )$ gives
\begin{align}
    \frac{d \bm\theta}{dt} &= - \frac{\eta}{\gamma} \sum_\mu \frac{\partial \ell_\mu}{\partial g_\mu } \frac{\partial f_\mu}{\partial \bm\theta} = - \eta_0 \gamma \sum_\mu \frac{\partial \ell_\mu}{\partial g_\mu } \frac{\partial f_\mu}{\partial \bm\theta} = O_{\gamma}(\gamma)
    \\
    \implies \frac{d}{dt} L &= \frac{1}{\gamma} \sum_\mu \frac{\partial \ell_\mu}{\partial g_\mu } \frac{\partial f_\mu}{\partial \bm\theta}  \frac{d \bm\theta}{dt} = - \eta_0  \sum_\mu \frac{\partial \ell_\mu}{\partial g_\mu } \frac{\partial \ell_\nu}{\partial g_\nu} \frac{\partial f_\mu }{\partial \bm\theta} \cdot \frac{\partial f_\nu}{\partial \bm\theta} = O_\gamma(1)
    \\
    \frac{d}{dt} \frac{\partial f(\x)}{\partial \bm\theta} &= \frac{\partial f^2}{\partial \bm\theta \partial \bm\theta} \cdot \frac{d \bm \theta}{dt} = O_\gamma(\gamma) 
\end{align}
We thus see an $O_\gamma(\gamma)$ increase in the relative rate of evolution of the parameter gradients compared to the loss $L$. This is very similar to the style of analysis in \cite{Chizat, Geiger_2020}.

\subsection{Linear networks with multiple outputs cannot specialize}
\label{section:no specialization in linear}
We derived a general expression for $K^{c,c'}(\bm{x},\bm{x'})$, defined in Eq.\ref{eq:def of subkernel}, for networks of any depth and show that it cannot show kernel specialization under an assumption of symmetry between target functions. 
Define $\bm{f}_l(\bm{x})=\bm{W^l} \bm{W^{l-1}}...\bm{W^1} \bm{x}$. Then
\begin{align}
    &N^{L+1} K^{c,c'}(\bm{x},\bm{x'})=\nabla_{\Theta}{r^{(c)}(\bm{x}})^T \nabla_{\Theta}{r^{(c')}(\bm{x'}}) \\
    =& \delta(c-c') \bm{f}_L(\bm{x})^T \bm{f}_L(\bm{x'}) \\
    &+ {\bm{V^c}}^T \bm{W^L} {\bm{W^L}}^T \bm{V^{c'}} \bm{f}_{L-1}(\bm{x})^T \bm{f}_{L-1}({\bm{x'}}) \\
    &+ {\bm{V^c}}^T \bm{W^L} \bm{W^{L-1}} {\bm{W^{L-1}}}^T {\bm{W^L}}^T \bm{V^{c'}} \bm{f}_{L-2}(\bm{x})^T \bm{f}_{L-2}(\bm{x'}) \\
    &+ {\bm{V^c}}^T \bm{W^L} \bm{W^{L-1}} \bm{W^{L-2}} {\bm{W^{L-2}}}^T {\bm{W^{L-1}}}^T {\bm{W^L}}^T \bm{V^{c'}} \bm{f}_{L-3}(\bm{x})^T \bm{f}_{L-3}(\bm{x'}) \\
    &+...
\end{align}
Defining scalar functions for $l<L$
\begin{equation}
\alpha_l(c,c') \equiv {\bm{V^c}}^T \bm{W^L} \bm{W^{L-1}}... \bm{W^{l+1}} {\bm{W^{l+1}}}^T... {\bm{W^{L-1}}}^T {\bm{W^L}}^T \bm{V^{c'}},
\end{equation}
one has
\begin{equation}
    N^{L+1} K^{c,c'}(\bm{x},\bm{x'}) = \delta(c-c') \bm{f}_L(\bm{x})^T \bm{f}_L(\bm{x'}) + \sum_{l=0}^{L-1} \alpha_l(c,c') \bm{f}_{l}(\bm{x})^T \bm{f}_{l}(\bm{x'}).
\end{equation}
It is thus a weighted sum of covariance of activations in all layers and the input. We make a class-symmetry ansatz that 

\begin{equation}
    \forall l,c,c': \alpha_l(c,c)=\alpha_l(c',c')=\tilde{\alpha_l}.
\end{equation}
To see why this ansatz is reasonable, define $\tilde{\bm{V}_l^c} \equiv {\bm{W^{l+1}}}^T... {\bm{W^{L-1}}}^T {\bm{W^L}}^T \bm{V^{c}}$. $\alpha_l(c,c)=\lVert  \tilde{\bm{V}_l^c} \rVert^2$; after learning, $y^{(c)}(\bm{x}) = r^{(c)}(\bm{x}) = N^{-(L+1)/2} ({\bm{W^{l+1}}}^T... {\bm{W^{L-1}}}^T {\bm{W^L}}^T \bm{V^{c}})^T \bm{f}_l(\bm{x})$. We then assume the covariance of $\bm{f}_l(\bm{x})$ projected along the direction of $\bm{V^{c}}$ to be approximately the same across $c$ and that $r^{(c)}(\bm{x})$ to have approximately the same variance. This would suggest that $\tilde{\bm{V}_l^c}$ should have the same norm across $c$.

Under the class-symmetry ansatz, 
\begin{equation}
    N^{L+1} K^{c,c}(\bm{x},\bm{x'}) =  \bm{f}_L(\bm{x})^T \bm{f}_L(\bm{x'}) + \sum_{l=0}^{L-1} \tilde{\alpha_l} \bm{f}_{l}(\bm{x})^T \bm{f}_{l}(\bm{x'})
\end{equation}
does not have $c$ dependence and thus cannot specialize.

\section{Theory of Kernel Alignment in Two-layer ReLU Networks}\label{app:relu_align_theory}

As in the main text,  we use the
following notation
\begin{equation}
\phi_{i}^{\mu}\equiv\phi\left(\bm{w_{i}}\cdot\bm{x^{\mu}}\right)
\end{equation}
\begin{equation}
D_{i}^{\mu}\equiv\frac{d}{dx}\phi\left(x\right)\Bigg|_{x=\bm{w_{i}}\cdot\bm{x^{\mu}}}.
\end{equation}
\begin{equation}
\bm{f}\in\mathbb{R}^{P}\quad f^{\mu}=f\left(\bm{x^{\mu}}\right)
\end{equation}
\subsection{A General Ansatz}
Throughout the derivations for the training dynamics of two-layer
ReLU networks, we make use of the following ansatz:
\begin{equation}
\forall\bm{x^{\mu}}\perp\bm{x^{\nu}},\forall t:\sum_{i}D_{i}^{\nu}\left(t\right)V_{i}\left(t\right)\phi_{i}^{\mu}\left(t\right)\approx\frac{1}{2}\sum_{i}V_{i}\left(t\right)\phi_{i}^{\mu}\left(t\right)=\frac{1}{2}f^{\mu}.\label{eq:bernoulli ansatz}
\end{equation}
While we do not prove this result rigorously, we provide a heuristic
argument for this. Since we assume $D_{i}^{\mu}\left(t\right)$ to
be static in time, they are determined by their initial values. Since
the weights are initialized as i.i.d. Gaussian, $\{D_{i}^{\mu}\}_{i=1,...,M}$
are random Bernoulli variables with mean of $1/2$. 

Given this, our ansatz will be true if $\{D_{i}^{\nu}\}_{i=1,...,M}$
and $\{V_{i}\left(t\right)\phi_{i}^{\mu}\left(t\right)\}_{i=1,...,M}$
are uncorrelated or weakly correlated. At initialization, this is
clearly true for $\bm{x^{\mu}}\perp\bm{x^{\nu}}$. However, the dynamics
of $V_{i}\left(t\right)$ depends on $\phi_{i}^{\nu}$, which has
the same sign as $D_{i}^{\nu}$. Therefore overtime, $\{V_{i}\}_{i=1,...,M}$
may generate correlation with $\{D_{i}^{\nu}\}_{i=1,...,M}$. However,
we speculate that when the dataset is isotropic ($\{\bm{x^{\nu}}\}_{\nu=1,..,P}$
are pairwise perpendicular and $P$ is large), this coupling is weak.
The dynamics are given by
\begin{equation}
\frac{dV_{i}}{dt}=\eta\sum_{\nu}\left(f^{\nu}-y^{\nu}\right)\phi_{i}^{\nu}.
\end{equation}
At initialization, $\{\left(f^{\nu}-y^{\nu}\right)\phi_{i}^{\nu}\}_{\nu=1,...P}$
are pairwise uncorrelated. Thus, $\{\phi_{i}^{\nu}\}$ has only a
$1/P$ effect on the dynamics of $V_{i}\left(t\right)$, which is
small for large $P$. If changes to the parameters are small (due
to large network width), this should hold approximately during training.

We numerically tested Eq.\ref{eq:bernoulli ansatz} in the various
settings and found excellent agreement. However, proving our conjecture rigorously is left for future work.

\subsection{Kernel Alignment in Two-layer ReLU Networks}
\label{appd: KA in two layer relu}
Assuming we are training a two layer ReLU network defined by

\begin{equation}
f^{\mu}=f\left(\bm{x^{\mu}}\right)=\sum_{i}V_{i}\phi\left(\bm{w_{i}}\cdot\bm{x^{\mu}}\right)
\end{equation}
on a training set with $P$ examples,  $\{\bm{x^{\mu}},y^{\mu}\}_{\mu=1,...,P}$
with a mean squared error loss.  

We assume $\{\bm{x^{\mu}}\}$ to be all unit vectors that are pairwise
perpendicular and $y^{\mu}\in\{-1,1\}$. As mentioned in the main
text, we assume $\{D_{i}^{\mu}\}$ to be static for all $\mu,i$ and
$\bm{y}^{T}\bm{f}>0$ at all times.

Batch gradient descent gives the following dynamics 
\begin{equation}
\frac{d\bm{w_{i}}}{dt}=-\eta\sum_{\nu}\left(f^{\nu}-y^{\nu}\right)V_{i}D_{i}^{\nu}\bm{x^{\nu}}=-\eta\sum_{\nu}\left(f^{\nu}-y^{\nu}\right)V_{i}D_{i}^{\nu}\bm{x^{\nu}}
\end{equation}

\begin{equation}
\frac{dV_{i}}{dt}=-\eta\sum_{\nu}\left(f^{\nu}-y^{\nu}\right)\phi{}_{i}^{\nu}
\end{equation}

\begin{align}
\frac{d\phi_{i}^{\mu}}{dt} & =\nabla_{\bm{w_{i}}}\phi_{i}^{\mu}\cdot\frac{d\bm{w_{i}}}{dt}=D_{i}^{\mu}\bm{x^{\mu}}\cdot\left[-\eta\sum_{\nu}\left(f^{\nu}-y^{\nu}\right)V_{i}D_{i}^{\nu}\bm{x^{\nu}}\right]\\
 & =D_{i}^{\mu}\bm{x^{\mu}}\cdot\left[-\eta\left(f^{\mu}-y^{\mu}\right)V_{i}D_{i}^{\mu}\bm{x^{\mu}}\right]\\
 & =\eta D_{i}^{\mu}\left(y^{\mu}-f^{\mu}\right)V_{i}.
\end{align}

As we did in the main text, we write the NTK as a sum of its two components

\begin{equation}
K_{\mu,\nu}=\nabla_{\bm{V}}f^{\mu}\cdot\nabla_{\bm{V}}f^{\nu}+\nabla_{\bm{W}}f^{\mu}\cdot\nabla_{\bm{W}}f^{\nu}\equiv(\bm{K}_{\bm{V}})_{\mu,\nu}+(\bm{K}_{\bm{W}})_{\mu,\nu}
\end{equation}

Define (for a unit vector $\bm{z}\in\mathbb{R}^{P}$) bilinear forms

\begin{align}
\alpha\left(\bm{z}\right) & =\bm{z}^{T}\bm{K}_{\bm{V}}\bm{z}=\sum_{\mu,\nu}z^{\mu}\nabla_{\bm{V}}f^{\mu}\cdot\nabla_{\bm{V}}f^{\nu}z^{\nu}=\sum_{i}\left(\sum_{\mu}z^{\mu}\phi_{i}^{\mu}\right)^{2}\\
\beta\left(\bm{z}\right) & =\bm{z}^{T}\bm{K}_{\bm{W}}\bm{z}=\sum_{\mu,\nu}z^{\mu}\nabla_{\bm{W}}f^{\mu}\cdot\nabla_{\bm{W}}f^{\nu}z^{\nu}\\
 & =\sum_{\mu,\nu}z^{\mu}\left[\sum_{i}V_{i}^{2}D_{i}^{\mu}D_{i}^{\nu}\left(\bm{x^{\mu}}\cdot\bm{x^{\nu}}\right)\right]z^{\nu}\\
 & =\sum_{\mu}v^{\mu2}\left[\sum_{i}V_{i}^{2}D_{i}^{\mu}\right]
\end{align}

Then

\begin{align}
\frac{d\alpha}{dt}\left(\bm{z}\right) & =\sum_{i}\sum_{\mu}\frac{d\alpha}{d\phi_{i}^{\mu}}\frac{d\phi_{i}^{\mu}}{dt}\\
 & =2\sum_{i}\left(\sum_{\nu}z^{\nu}\phi_{i}^{\nu}\right)\sum_{\mu}z^{\mu}\eta D_{i}^{\mu}\left(y^{\mu}-f^{\mu}\right)V_{i}\\
 & =2\eta\sum_{\mu}\left(y^{\mu}-f^{\mu}\right)z^{\mu}\sum_{\nu}z^{\nu}\left(\sum_{i}\phi_{i}^{\nu}D_{i}^{\mu}V_{i}\right)
\end{align}

We now consider the object $\sum_{i}\phi_{i}^{\nu}D_{i}^{\mu}V_{i}.$
For $\mu=\nu$, $\sum_{i}\phi_{i}^{\nu}D_{i}^{\nu}V_{i}=f^{\nu}.$
For $\mu\neq\nu$, we use the ansatz in Eq.\ref{eq:bernoulli ansatz}.
Hence

\begin{align}
\frac{d\alpha}{dt}\left(\bm{z}\right) & \approx\eta\sum_{\mu}\left(y^{\mu}-f^{\mu}\right)z^{\mu}z^{\mu}f^{\mu}+\eta\sum_{\mu}\left(y^{\mu}-f^{\mu}\right)z^{\mu}\sum_{\nu\neq\mu}z^{\nu}f^{\nu}\\
 & =\eta\sum_{\mu}\left(y^{\mu}-f^{\mu}\right)z^{\mu}z^{\mu}f^{\mu}+\eta\left[\left(\bm{y}-\bm{f}\right)^{T}\bm{z}\right]\left[\bm{z}^{T}\bm{f}\right]
\end{align}

To track the anisotropy of $\bm{K}_{\bm{V}}$, we first consider its
average over all unit vectors with a uniform measure

\begin{align}
\left\langle \frac{d\alpha}{dt}\left(\bm{z}\right)\right\rangle _{\bm{z}} & =\frac{1}{P}\eta\left(\bm{y}-\bm{f}\right)^{T}\bm{f}+\frac{1}{P}\eta\left[\left(\bm{y}-\bm{f}\right)^{T}\bm{f}\right]\\
 & =\frac{1}{P}2\eta\left(\bm{y}-\bm{f}\right)^{T}\bm{f}\\
 & \approx\frac{1}{P}2\eta\bm{y}^{T}\bm{f}
\end{align}

we then compute it for a unit vector in the direction of $\bm{y}$,
$\bm{y}/\sqrt{P}$

\begin{align}
\frac{d\alpha}{dt}\left(\bm{y}/\sqrt{P}\right) & =\frac{1}{P}\eta\sum_{\mu}\left(y^{\mu}-f^{\mu}\right)f^{\mu}+\frac{1}{P}\eta\left[\left(\bm{y}-\bm{f}\right)^{T}\bm{y}\right]\left[\bm{y}^{T}\bm{f}\right]\\
 & \approx\frac{1}{P}\eta\bm{y}^{T}\bm{f}+\eta\bm{y}^{T}\bm{f}\approx\eta\bm{y}^{T}\bm{f}.
\end{align}

Hence, for $\bm{y}^{T}\bm{f}>0$ and large $P$, we have

\begin{equation}
\frac{d\alpha}{dt}\left(\bm{y}/\sqrt{P}\right)\gg\left\langle \frac{d\alpha}{dt}\left(\bm{z}\right)\right\rangle _{\bm{z}}.
\end{equation}

This show that one part of the NTK, $\bm{K}_{\bm{V}}$, is growing
in the direction of $\bm{y}$ and static or contracting in directions
perpendicular to $\bm{y}$, thus aligning to $\bm{y}$ over time.
We now analyze the anisotropy of $\bm{K_{W}}$.

Now for $\beta$, since we assume $\{\bm{D}^{\mu}\}$ to be static,
its dynamics are entirely driven by $\bm{V}$.

\begin{equation}
\frac{d\beta}{dV_{i}}\left(\bm{z}\right)=2\sum_{\mu}z_{\mu}^{2}D_{i}^{\mu}V_{i}
\end{equation}

\begin{align}
\frac{d\beta}{dt}\left(\bm{z}\right) & =\sum_{i}\frac{d\beta}{dV_{i}}\frac{dV_{i}}{dt}=2\eta\sum_{i}\sum_{\mu}z_{\mu}^{2}D_{i}^{\mu}V_{i}\sum_{\nu}\left(y^{\nu}-f^{\nu}\right)\phi{}_{i}^{\nu}\\
 & =2\eta\sum_{\mu}z_{\mu}^{2}\sum_{\nu}\left(y^{\nu}-f^{\nu}\right)\sum_{i}D_{i}^{\mu}V_{i}\phi{}_{i}^{\nu}
\end{align}

Analysis of the object $\sum_{i}D_{i}^{\mu}V_{i}\phi{}_{i}^{\nu}$
is the same as above, resulting in 

\begin{equation}
\frac{d\beta}{dt}\left(\bm{z}\right)\approx\eta\left\Vert \bm{z}\right\Vert ^{2}\left(\bm{y}-\bm{f}\right)^{T}\bm{f}.
\end{equation}

\subsection{Kernel Specialization in Two-layer ReLU Networks}
\label{apd:specialization}
We consider a two-layer ReLU network with $C$ output heads. The output
of the $c$th head is

\begin{equation}
f_{c}^{\mu}\equiv f_{c}\left(\bm{x^{\mu}}\right)=\sum_{i}V_{i}^{c}\phi\left(\bm{w_{i}}\cdot\bm{x^{\mu}}\right)
\end{equation}

Assume we are classifying a mixture of $C$ Gaussians, $\{\mathcal{N}(\bm{\mu}_{c},\sigma^{2}\mathbb{I}\}_{c=1,...,C}$.
$P/C$ points are drawn from each Gaussian. Centers of the Gaussians,
$\{\bm{\mu}_{c}\}$, are all unit vectors and pairwise perpendicular.
Then data points from class $c$ can be written as

\begin{equation}
\bm{x}^{\mu\in\text{class c}}=\bm{\mu}_{c}+\delta\bm{x}^{\mu}.
\end{equation}

This leads to the following input correlation structure

\begin{equation}
\bm{x}^{\mu}\cdot\bm{x}^{\nu}=\begin{cases}
1 & \text{if}\mu=\nu\\
m & \text{if}\mu\neq\nu\in\text{same class}\\
0 & o.w.
\end{cases}
\end{equation}

Assume $\sigma^{2}$ to be sufficiently small that

\begin{equation}
\phi\left(\bm{w}_{i}\cdot\bm{\mu}_{c}+\delta\bm{x}^{\mu}\right)\approx\phi\left(\bm{w}_{i}\cdot\bm{\mu}_{c}\right)+\left(\bm{w}_{i}\cdot\delta\bm{x}^{\mu}\right)\phi'\left(\bm{w}_{i}\cdot\bm{\mu}_{c}\right)
\end{equation}

\begin{equation}
\phi'\left(\bm{w}_{i}\cdot\bm{\mu}_{c}+\delta\bm{x}^{\mu}\right)\approx\phi'\left(\bm{w}_{i}\cdot\bm{\mu}_{c}\right)+\left(\bm{w}_{i}\cdot\delta\bm{x}^{\mu}\right)\phi''\left(\bm{w}_{i}\cdot\bm{\mu}_{c}\right)
\end{equation}

If $P/C$ is sufficiently large, then

\begin{equation}
\sum_{\mu\in\text{class c}}\phi_{i}^{\mu}\approx P/C\phi_{i}^{c},\quad\phi_{i}^{c}\equiv\phi\left(\bm{w}_{i}\cdot\bm{\mu}_{c}\right)
\end{equation}

\begin{equation}
\sum_{\mu\in\text{class c}}D_{i}^{\mu}\approx P/CD_{i}^{c},\quad D_{i}^{c}\equiv\phi'\left(\bm{w}_{i}\cdot\bm{\mu}_{c}\right).
\end{equation}

It follows that $f_{c}\left(\bm{\mu}_{c}\right)\approx\sum_{\mu\in\text{class c}}f_{c}^{\mu}.$
Since we assume $\sum_{\mu}f_{c}\left(\bm{x^{\mu}}\right)y_{c}^{\mu}>0$
and $y_{c}^{\mu\in\text{class c}}=1$, $f_{c}\left(\bm{\mu}_{c}\right)>0$.

We would like to analyze the anisotropy in the dynamics of $c$th
subkernel, defined as

\begin{align}
\bm{K^{cc}} & =\sum_{d}\nabla_{\bm{V^{d}}}f_{c}^{\mu}\cdot\nabla_{\bm{V^{d}}}f_{c}^{\nu}+\nabla_{\bm{W}}f_{c}^{\mu}\cdot\nabla_{\bm{W}}f_{c}^{\nu}=\nabla_{\bm{V^{c}}}f_{c}^{\mu}\cdot\nabla_{\bm{V^{c}}}f_{c}^{\nu}+\nabla_{\bm{W}}f_{c}^{\mu}\cdot\nabla_{\bm{W}}f_{c}^{\nu}\\
 & \equiv\bm{K_{v}^{cc}}+\bm{K_{W}^{cc}}.
\end{align}

In particular, we want to show that it grows in the direction of $\bm{y_{c}}$
at a rate faster than in the directions of $\bm{y_{d\neq c}}$. As
argued in the main text, $\bm{K_{v}^{cc}}$ cannot be anisotropic
for different $\bm{y_{d}}$. Thus, we only track the anisotropic dynamics
of $\bm{K_{W}^{cc}}$ with the bilinear form (all $\{\bm{y_{d}}\}_{d=1,...,C}$
have the same norm)

\begin{align}
\text{\ensuremath{B_{\bm{W}}^{c}}(\ensuremath{\bm{y_{d}}})} & =\bm{y_{d}}^{T}\bm{K_{W}^{cc}}\bm{y_{d}}=\sum_{\mu,\nu}^{P}y_{d}^{\mu}y_{d}^{\nu}\bm{x}_{\mu}^{T}\bm{x}_{\nu}\sum_{i}D_{i}^{\mu}D_{i}^{\nu}(\bm{V}_{c})_{i}^{2}.\\
 & =\sum_{\mu\in\text{class d}}\sum_{i}\left(D_{i}^{\mu}\right)^{2}(\bm{V}_{c})_{i}^{2}+m\sum_{k}\sum_{\mu\neq\nu\text{\ensuremath{\in\text{class d}}}}^{P/C}\sum_{i}D_{i}^{\mu}D_{i}^{\nu}(\bm{V}_{c})_{i}^{2}\\
 & =\sum_{\mu\in\text{class d}}\sum_{i}D_{i}^{\mu}(\bm{V}_{c})_{i}^{2}+m\sum_{k}\sum_{\mu\neq\nu\text{\ensuremath{\in\text{class d}}}}^{P/C}\sum_{i}D_{i}^{\mu}D_{i}^{\nu}(\bm{V}_{c})_{i}^{2}\quad\text{because }\text{\ensuremath{D_{i}^{\mu}}\ensuremath{\ensuremath{\in}\{0,1\}}}
\end{align}

The gradient descent dynamics of $\bm{V_{c}}$ is

\begin{equation}
\frac{d(\bm{V}_{c})_{i}}{dt}=-\eta\sum_{\mu}\left(f_{c}^{\mu}-y_{c}^{\mu}\right)\phi_{i}^{\mu}\approx\eta\sum_{\mu\in\text{class c}}\phi_{i}^{\mu}.
\end{equation}

Then dynamics of $\text{\ensuremath{B_{\bm{W}}^{c}}(\ensuremath{\bm{y_{d}}})}$
follow

\begin{align}
\eta^{-1}\frac{dB_{\bm{W}}^{c}(\bm{y_{d}})}{dt} & =\eta^{-1}\sum_{i}\frac{dB_{\bm{W}}^{c}(\bm{y_{d}})}{d(\bm{V}_{c})_{i}}\frac{d(\bm{V}_{c})_{i}}{dt}=\sum_{i}(\bm{V}_{c})_{i}\left[\sum_{\mu\in\text{class d}}D_{i}^{\mu}+m\sum_{\mu\neq\nu\text{\ensuremath{\in\text{class d}}}}^{P/C}D_{i}^{\mu}D_{i}^{\nu}\right]\sum_{\omega\in\text{class c}}\phi_{i}^{\omega}\\
 & \approx\frac{P}{C}\sum_{i}\left[\frac{P}{C}D_{i}^{d}(\bm{V}_{c})_{i}+m\sum_{\mu\neq\nu\text{\ensuremath{\in\text{class d}}}}^{P/C}D_{i}^{d}(\bm{V}_{c})_{i}\right]\phi_{i}^{c}\\
 & =\frac{P}{C}\left[\frac{P}{C}+mP^{2}/C^{2}\right]\sum_{i}D_{i}^{d}(\bm{V}_{c})_{i}\phi_{i}^{c}
\end{align}

For $B_{\bm{W}}^{c}(\bm{y_{c}})$, this simplifies to $\frac{P}{C}\left[\frac{P}{C}+mP^{2}/C^{2}\right]f_{c}\left(\bm{\mu}_{c}\right)>0.$
On the other hand, for $B_{\bm{W}}^{c}(\bm{y_{d\neq c}}),$ one has
$\sum_{i}D_{i}^{d}(\bm{V}_{c})_{i}\phi_{i}^{c}\approx\frac{1}{2}f_{c}\left(\bm{\mu}_{c}\right)$(following
Eq.\ref{eq:bernoulli ansatz}) and thus

\begin{equation}
\eta^{-1}\frac{dB_{\bm{W}}^{c}(\bm{y_{d}})}{dt}\approx\frac{1}{2}\frac{P}{C}\left[\frac{P}{C}+mP^{2}/C^{2}\right]f_{c}\left(\bm{\mu}_{c}\right)>0
\end{equation}

\section{Experimental Details}
\label{expt_details}
We train our models on a Google Colab GPU and include code to reproduce all experimental results in the supplementary materials. To match our theory, we use fixed learning rate SGD. Both evaluation of the infinite width kernels and training were performed with the Neural Tangents API \cite{neuraltangents2020}. 

\subsection{Wide Res-Net with CIFAR-10}
\label{appd:wide resnet details}
We used the Neural Tangents API implementation of the Wide ResNet model which can be found on the ReadME of the github \url{https://github.com/google/neural-tangents#infinitely-wideresnet}. We used a width factor of $k=3$ and a blocksize of $b=2$. The only change between our experiments was rescaling the output of the network $g(\x) = \frac{1}{\gamma} f(\x)$. We used $100$ CIFAR-10 images taken at random from the first two classes and used binary labels $y_\mu \in \{\pm 1\}$.  

\subsection{ReLU networks trained on MNIST}
\label{subsub: two layre relu on mnist}
The network has 2 hidden layers with hidden dimension of 500. The network was trained with full-batch gradient descent on a subset of the MNIST dataset (1000 examples). The target functions are "10-hot" vectors encoding the classes. The learning rate was fixed at 20 (the loss function is averaged over the batch and the 10 classes).

These simulations were performed in Google Colab using GPU acceleration and \texttt{jax}.

\subsection{ReLU convolutional networks trained on CIFAR-10}
\label{subsub: two layre relu on cifar10}
The network has two convolutional layers, each with 33 channels and each filter is $7 \times 7$. The target functions are "10-hot" vectors encoding the classes. The network was trained on 9000 random samples from the CIFAR dataset with a batchsize of 100 and the mean squared error loss function. The learning rate was fixed at 20 (the loss function is averaged over the batch and the 10 classes). To save computational resources, the kernel was estimated using Gram matrices computed on 300 examples.

These simulations were performed in Google Colab using GPU acceleration and \texttt{jax}.

\subsection{Two-layer ReLU networks trained on random binary classification.}
\label{subsub: two layer relu random binary details}
We used input dimension ($N$) 1000, hidden dimension ($M$) 2000 and the number of examples ($P$) 100. The network has one hidden layer. The network did not have biases. The first layer weights $\bm W$ were initialized as i.i.d. Gaussian with variance $1/N$, and the second layer weights $\bm V$ where initialized as i.i.d. Gaussian with variance $1/M$. 

The input vectors were sampled from an $N$-dimensional Gaussian distribution $\mathcal{N}(\bm 0, 1/\sqrt{N} \mathbb{I})$ and the labels were random binary labels $y^\mu \in {-1,1}$ with equal probability. 

Full-batch gradient descent with a fixed learning rate (0.1) was performed on $\bm{W,V}$ using the mean squared error loss function $L=P^{-1} \sum_\mu (y^\mu - f^\mu)^2$. These simulations were performed on a personal computer using \texttt{pytorch} and no GPU.

\subsection{Two-layer ReLU networks trained on classifying a mixture of Gaussians.}
\label{subsub: two layer relu mixture}
We used input dimension ($N$) 1000, hidden dimension ($M$) 2000 and the number of examples ($P$) 500. The network has one hidden layer. There are 10 classes ($C$). The target function for each output node is an one-hot encoding vector. The network did not have biases. The first layer weights $\bm W$ were initialized as i.i.d. Gaussian with variance $1/N$, and the second layer weights $\bm V$ where initialized as i.i.d. Gaussian with variance $1/M$. 

The centers of the Gaussians are randomly drawn from the $N$-dimensional unit sphere and the variance of all Gaussians ($\sigma^2$) is 0.01. The network has 10 output nodes. The classes are balanced such that for each class there are $P/C=50$ data points in the training set.

Full-batch gradient descent with a fixed learning rate (0.2) was performed on $\bm{W,V}$ using the mean squared error loss function $L=P^{-1} \sum_\mu \sum_c^C (y^\mu_c - f^\mu_c)^2$. These simulations were performed on a personal computer using \texttt{pytorch} and no GPU.
%%%%%%%%%%%%%%%%%%%%%%%%%%%%%%%%%%%%%%%%%%%%%%%%%%%%%%%%%%%%%%%%%%%%%%%%%%%%%%%
%%%%%%%%%%%%%%%%%%%%%%%%%%%%%%%%%%%%%%%%%%%%%%%%%%%%%%%%%%%%%%%%%%%%%%%%%%%%%%%

\end{document}